\documentclass[10pt,twocolumn,letterpaper]{article}

\usepackage{iccv}
\usepackage{times}
\usepackage{epsfig}
\usepackage{graphicx}
\usepackage{amsmath}
\usepackage{amssymb}
\usepackage{subfigure} 
\usepackage{float} 
\usepackage{CJK} 
\usepackage{amsmath} 
\usepackage{amssymb} 
\usepackage{longtable} 
\usepackage{multirow} 
\usepackage{array} 
\usepackage{chngpage} 
\usepackage{mathrsfs} 

\usepackage[pagebackref=true,breaklinks=true,letterpaper=true,colorlinks,bookmarks=false]{hyperref}

\iccvfinalcopy 

\ificcvfinal\fi
\begin{document}

\title{Deep Direct Regression for Multi-Oriented Scene Text Detection}

\author{
	{Wenhao He\quad Xu-Yao Zhang\quad Fei Yin\quad Cheng-Lin Liu} \\
	National Laboratory of Pattern Recognition (NLPR) \\
	Institute of Automation, Chinese Academy of Sciences, Beijing, China \\
	\tt\small Email: \{wenhao.he, xyz, fyin, liucl\}@nlpr.ia.ac.cn
}

\maketitle

\begin{abstract}
	In this paper, we first provide a new perspective to divide existing high performance object detection methods into direct and indirect regressions. Direct regression performs boundary regression by predicting the offsets from a given point, while indirect regression predicts the offsets from some bounding box proposals.
	Then we analyze the drawbacks of the indirect regression, which the recent state-of-the-art detection structures like Faster-RCNN and SSD follows, for multi-oriented scene text detection, and point out the potential superiority of direct regression.
	To verify this point of view, we propose a deep direct regression based method for multi-oriented scene text detection.
	Our detection framework is simple and effective with a fully convolutional network and one-step post processing.
	The fully convolutional network is optimized in an end-to-end way and has bi-task outputs where one is pixel-wise classification between text and non-text, and the other is direct regression to determine the vertex coordinates of quadrilateral text boundaries.
	The proposed method is particularly beneficial for localizing incidental scene texts.
	On the ICDAR2015 Incidental Scene Text benchmark, our method achieves the F1-measure of 81\%, which is a new state-of-the-art and significantly outperforms previous approaches.
	On other standard datasets with focused scene texts, our method also reaches the state-of-the-art performance.
\end{abstract}

\section{Introduction}
\label{Sec.1}
	Scene text detection has drawn great interests from both computer vision and machine learning communities because of its great value in practical uses and the technical challenges.
	Owing to the significant achievements of deep convolutional neural network (CNN) based generic object detection in recent years, scene text detection also has been greatly improved by regarding text words or lines as objects.
	High performance methods for object detection like Faster-RCNN \cite{faster-rcnn}, SSD \cite{ssd} and YOLO \cite{yolo} have been modified to detect horizontal scene texts \cite{deep-text} \cite{yolo-text} \cite{ctpn} \cite{textbox} and gained great improvements.
	However, for multi-oriented text detection, methods like Faster-RCNN and SSD which work well for object and horizontal text detection may not be good choices. To illustrate the reasons, first we explain the definitions of indirect and direct regression in detection task.
	
	
	\begin{figure}
		\label{Fig.direct_indirect_regression}
		\centering
		\subfigure[]{
			\includegraphics[scale=0.14]{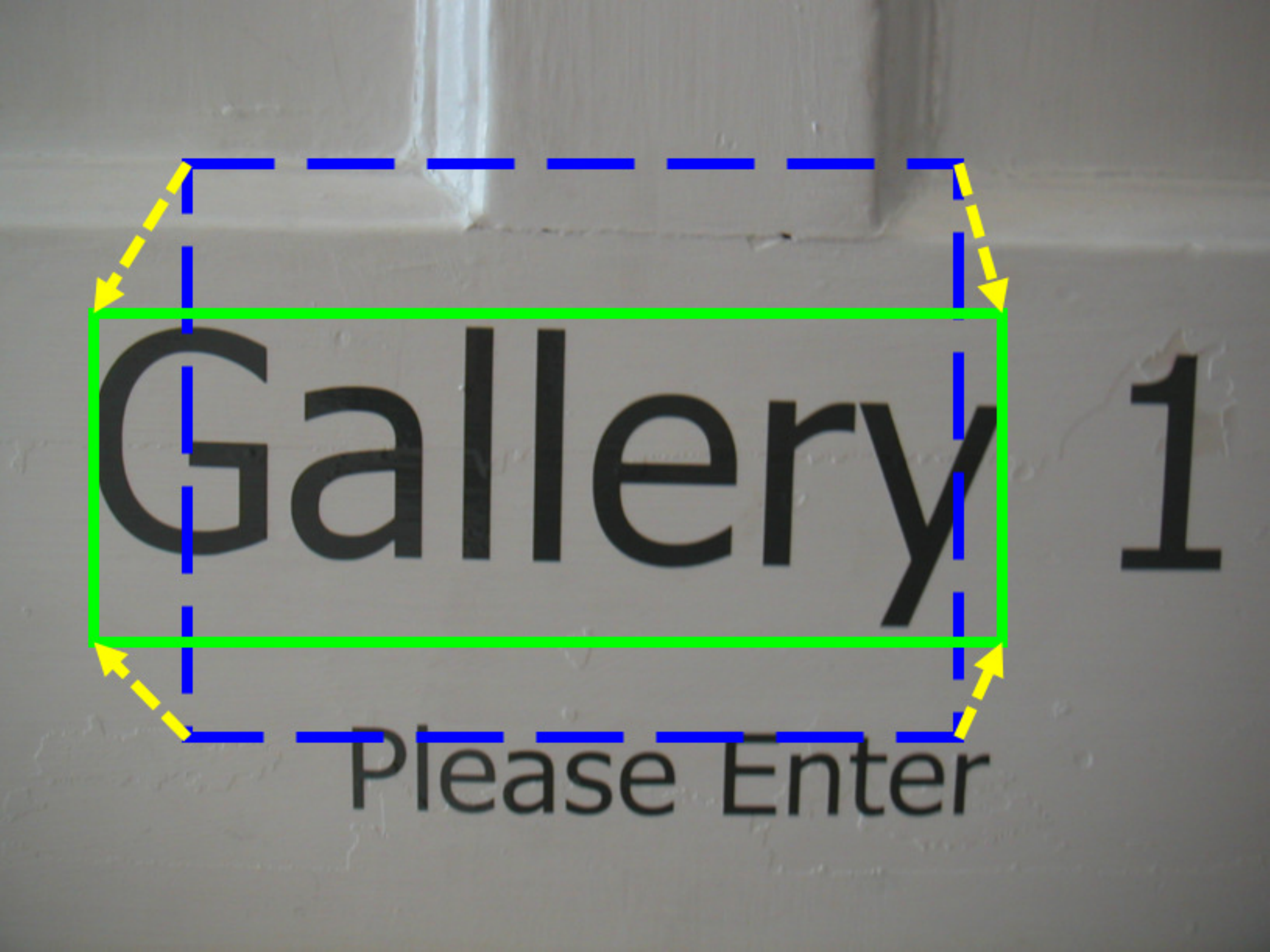}}
		\subfigure[]{
			\includegraphics[scale=0.14]{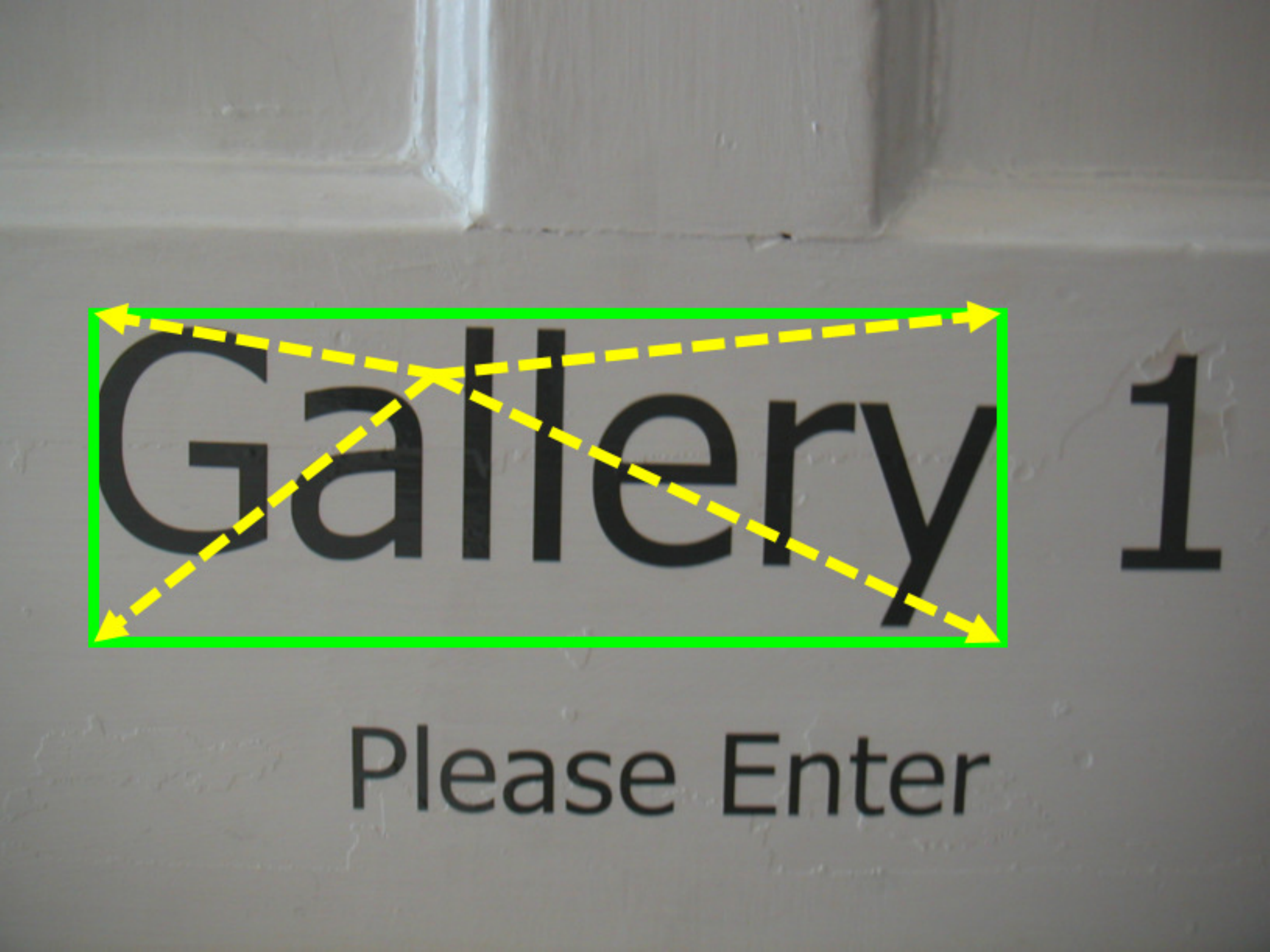}}
		\caption{Visualized explanation of indirect and direct regression. The solid green lines are boundaries of text ``Gallery'', the dash blue lines are boundaries of text proposal, and the dashed yellow vectors are the ground truths of regression task. (a) The indirect regression predicts the offsets from a proposal. (b) The direct regression predicts the offsets from a point.}
	\end{figure}
	
	
	
	\noindent \textbf{Indirect Regression.} \ For most CNN based detection methods like Fast-RCNN \cite{fast-rcnn}, Faster-RCNN, SSD, Multi-Box \cite{multibox}, the regression task is trained to regress the offset values from a proposal to the corresponding ground truth (See Fig.\hyperref[Fig.direct_indirect_regression]{1}.a). We call these kinds of approaches indirect regression. 
	
	\noindent \textbf{Direct Regression.} \ For direct regression based methods, the regression task directly outputs values corresponding with the position and size of an object from a given point (See Fig.\hyperref[Fig.direct_indirect_regression]{1}.b). Take DenseBox \cite{densebox} as an instance, this model learns to directly predict offsets from bounding box vertexes to points in region of interest.
	
	\begin{figure}
		\label{Fig.anchor_not_suitable}
		\centering
		\includegraphics[scale=0.14]{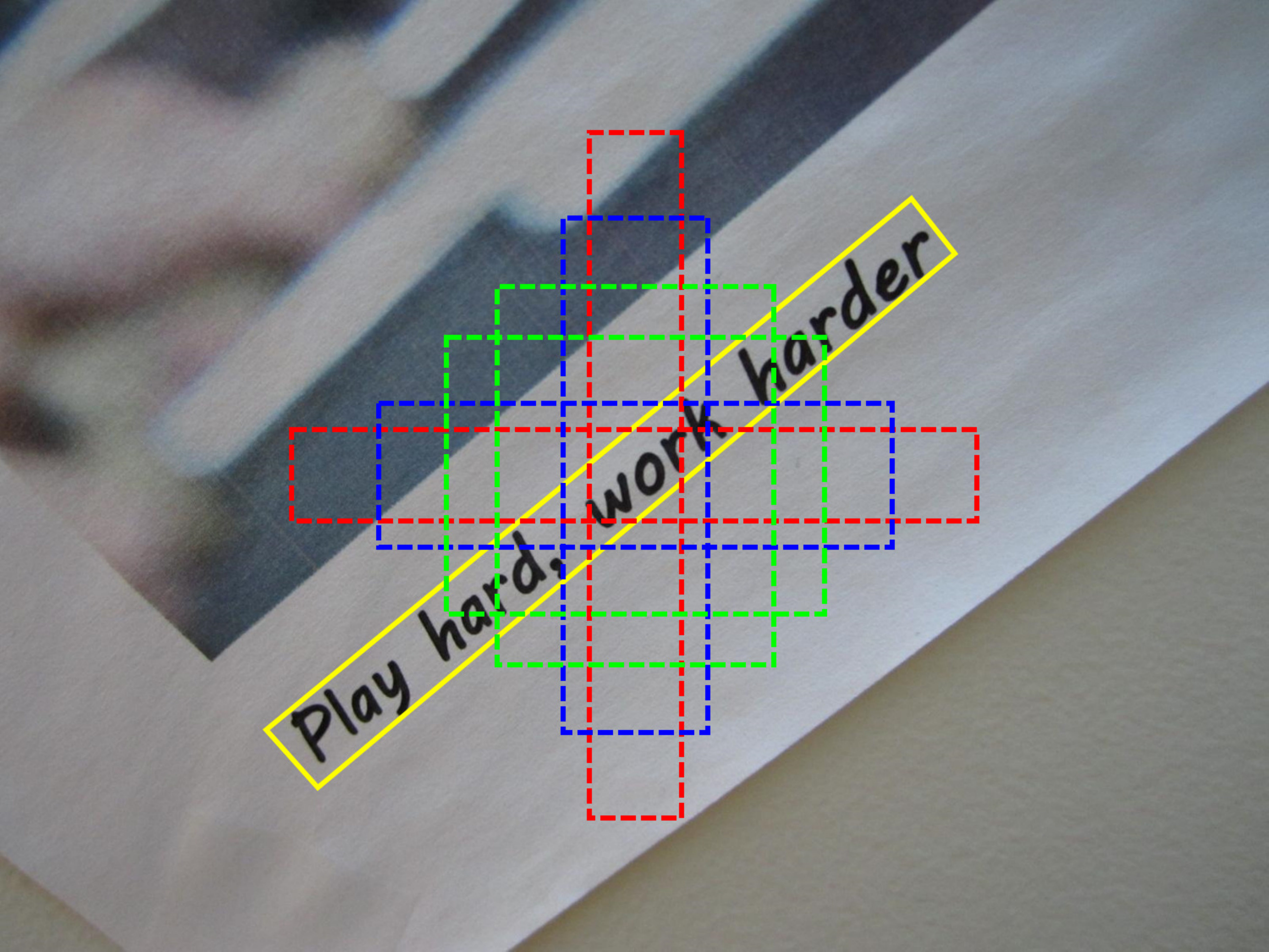}
		\caption{Illustration for the deficiency of anchor mechanism in detecting long and heavily inclined text words or lines. The solid yellow lines are boundaries of the text line and the dashed lines are boundaries of anchors. There is no anchor that has sufficient overlap with the text line in this image.}
		
	\end{figure}
	
	Indirect regression based detection methods may not be effective for multi-oriented text detection, even methods like Faster-RCNN and SSD have reached state-of-the-art performance for object detection and are also implemented for horizontal scene text detection.
	The reasons are mainly in three folds.
	First, there are few robust methods to generate word-level or line-level proposals for multi-oriented text. Most previous methods could only provide proposals of character-level by extracting connected components.
	Second, anchor mechanism in Faster-RCNN may not be an effective solution to generate text proposals. The anchor mechanism can be deemed as rectangular proposals of various sizes and aspect ratios being evenly placed on an image, and setting proposals which have high overlap with ground truths as positive, otherwise as ``NOT CARE'' or negative. However, for multi-oriented scene texts which are long and heavily inclined, there may be no proper anchor that has sufficient overlap with them as shown in Fig.\hyperref[Fig.anchor_not_suitable]{2}.
	Third, adopting anchor mechanism may cause the detection system less efficient. Taking horizontal scene text detection as instance, unlike generic objects, horizontal scene texts tend to have larger variation in sizes and aspect ratios, which requires more complicate design of anchors. The anchors used in \cite{deep-text} \cite{textbox} are much more than traditional Faster-RCNN in both scale and aspect ratio. As to multi-oriented text detection, inclined text proposals may be generated by adopting multi-oriented anchors like \cite{dmpn}, however, this will cost much more running time in the meanwhile and the proposal may not be an optimal choice. Based on the analysis above, direct regression based methods which need no proposals beforehand could be a better choice to produce the irregular quadrilateral boundaries for multi-oriented scene texts.
	
	In this paper, we propose a novel multi-oriented text detection method based on direct regression. Our method is particularly beneficial for localizing quadrilateral boundaries of incidental scene texts which are hard to identify the constitute characters and have large variations in scales and perspective distortions. 
	On the ICDAR2015 Incidental Scene Text benchmark, we obtain F1-measure of 81\%, which is a new state-of-the-art and surpass the second placed method by a large margin. On other popular datasets of focused images, the proposed method also reaches the state-of-the-art performance.
	
	The proposed method has several novelties and advantages.
	First, this is the first direct regression based method for multi-oriented scene text detection.
	Second, the whole pipeline of the proposed method only has two parts in which one is a convolutional neural network and the other is a one-step post processing call Recalled Non-Maximum Suppression. Modules like line grouping and word partition are removed which saves much effort on tuning parameters.
	Third, since our method could predict irregular quadrilateral boundaries, it has great superiority in incidental texts detection task which needs to localize four vertexes of each word-level text.

	The rest of this paper is organized as follows: In Section \hyperref[Sec.2]{2} we give a brief review of scene text detection and generic object detection, in Section \hyperref[Sec.3]{3} we introduce details of our proposed method, in Section \hyperref[Sec.4]{4} we present the results on benchmarks and the rationality analysis of the performance, as well as comparisons to other scene text detection systems, and in Section \hyperref[Sec.5]{5} we conclude this paper.

\section{Related Work}
\label{Sec.2}
	\noindent{\textbf{Scene Text Detection.}} \ Most scene text detection methods \cite{fcn-text} \cite{ctpn} \cite{huang2013text} \cite{mser1} \cite{yfpan} treat text as a composite of characters, so they first localize character or components candidates and then group them into a word or text line. 
	Even for multi-oriented text, methods like \cite{yao2012detecting} \cite{yin2015multi} \cite{kang2014orientation} also follow the same strategy and the multi-oriented line grouping is accomplished by either rule based methods or more complex graphic model.
	However, for texts in the ICDAR2015 Incidental Scene Text Dataset \cite{karatzas2015icdar}, some blurred or low resolution characters in a word could not be well extracted, which hinders the performance of localization.
	
	Recently, some text detection methods discard the text composition and take text words or lines as generic objects.
	The method in \cite{symmetry-text} makes use of the symmetric feature of text lines and tries to detect text line as a whole. 
	Despite the novelty of this work, the feature it uses is not robust for cluttered images.
	The method in \cite{yolo-text} adopts the framework for object detection in \cite{yolo}, but the post-processing relies on the text sequentiality.
	The methods in \cite{deep-text} and \cite{textbox} are based on Faster-RCNN \cite{faster-rcnn} and SSD \cite{ssd} respectively. They both attempt to convert text detection into object detection and the performance on horizontal text detection demonstrate their effectiveness. However, constrained by the deficiency of indirect regression, those two methods may not be suitable for multi-oriented scene text detection. 
	The method in \cite{dmpn} rotates the anchors into more orientations and tries to find the best proposal to match the multi-oriented text. Deficiency of this method is that the best matched proposal may not be an optimal choice since the boundary shape of scene texts is arbitrary quadrilateral while the proposal shape is parallelogram.
	
	\noindent{\textbf{Generic Object Detection.}} \ Most generic object detection frameworks are multi-task structure with a classifier for recognition and a regressor for localization.
	According to the distinction of regressor, we divide these methods into direct and indirect regression.
	The direct regression based methods like \cite{densebox} predict size and localization of objects straightforwardly.
	The indirect regression based methods like \cite{fast-rcnn} \cite{faster-rcnn} \cite{multibox} \cite{ssd} predict the offset from proposals to the corresponding ground truths.
	It should be noted that, the proposals here can be generated by either class-agnostic object detection methods like \cite{uijlings2013selective} or simple clustering \cite{multibox}, as well as anchor mechanism \cite{faster-rcnn} \cite{ssd}.
	
	Although most of the recent state-of-the-art approaches are indirect regression based methods, considering the wide variety of texts in scale, orientation, perspective distortion and aspect ratio, direct regression might have the potential advantage of avoiding the difficulty in proposal generation for multi-oriented texts. This is the main contribution of this paper.

\section{Proposed Methodology}
\label{Sec.3}
\begin{figure*}
	\label{Fig.whole_pipeline}
	\centering
	\includegraphics[scale=0.26]{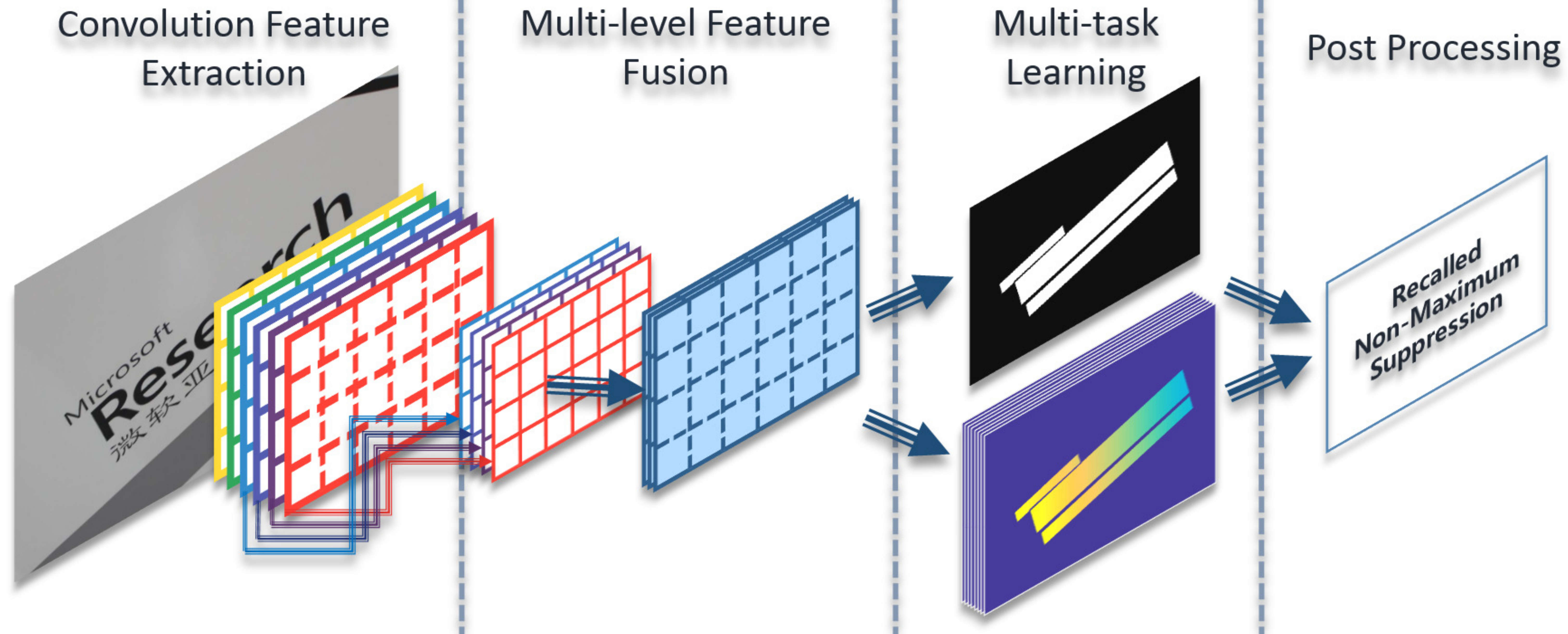}
	\caption{Overview of the proposed text detection method.}
\end{figure*}

The proposed detection system is diagrammed in Fig.\hyperref[Fig.whole_pipeline]{3}. It consists of four major parts: the first three modules, namely convolutional feature extraction, multi-level feature fusion, multi-task learning, together constitute the network part, and the last post processing part performs recalled NMS, which is an extension of traditional NMS.

\subsection{Network Architecture}
\label{Sec.3.1}
The convolutional feature extraction part is designed so that the maximum receptive field is larger than the input image size $S$.
This ensures the regression task could see long texts and give more accurate boundary prediction.
Considering that the text feature is not as complicated as that of generic objects, our network tends to employ less parameters than models designed for ImageNet to save computation.

The feature fusion part referring to the design in \cite{fcn} combine convolutional features from four streams to capture texts of multiple scales.
However, to reduce computation, we only up-sample the fused feature to quarter size of the input image.

The multi-task part has two branches.
The classification task output $\mathcal{M}_{cls}$ is a $\frac{S}{4} \times \frac{S}{4}$ 2nd-order tensor and it can be approximated as down-sampled segmentation between text and non-text for input images. Elements in $\mathcal{M}_{cls}$ with higher score are more likely to be text, otherwise non-text;
The regression task output $\mathcal{M}_{loc}$ is a $\frac{S}{4} \times \frac{S}{4} \times 8$ 3rd-order tensor.
The channel size of $\mathcal{M}_{loc}$ indicates that we intend to output 8 coordinates, corresponding to the quadrilateral vertexes of the text. 
The value at $\left(w, h, c\right)$ in $\mathcal{M}_{loc}$ is denoted as $L_{\left(w, h, c\right)}$, which means the offset from coordinate of a quadrilateral vertex to that of the point at $\left(4w, 4h\right)$ in input image, and therefore, the quadrilateral $\mathcal{B} \left(w, h\right)$ can be formulated as 

\begin{equation}
\begin{split}
& \mathcal{B} \left(w, h\right) = \\ & \left\{ L_{\left(w, h, 2n\! -\! 1\right)}\! +\! 4w, L_{\left(w, h, 2n\right)}\! +\! 4h \big| n \in \left\{1,2,3,4\right\}\right\}
\end{split}
\end{equation}

\begin{figure}
	\label{Fig.network_structure}
	\centering
	\includegraphics[scale=0.12]{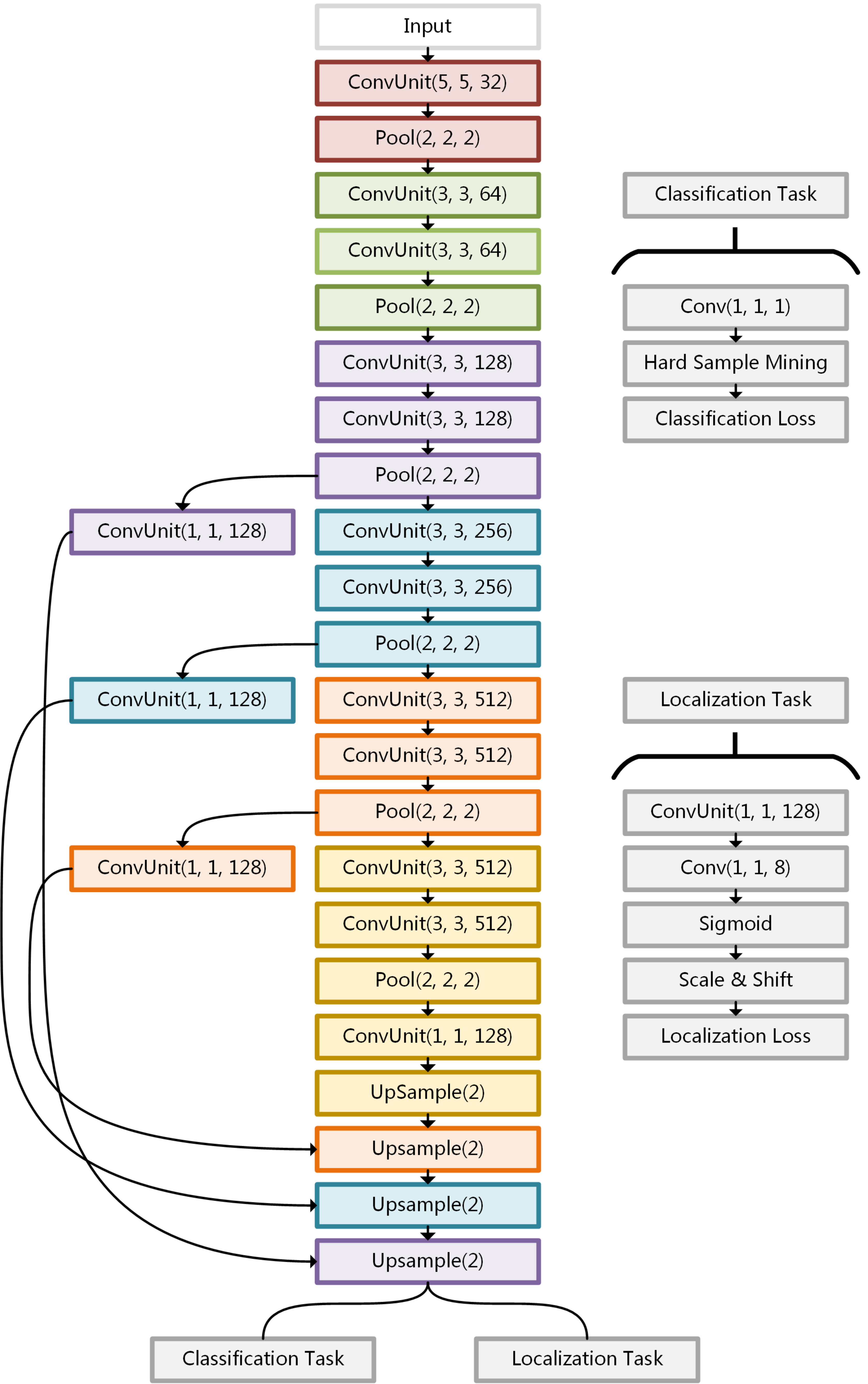}
	\caption{Structure of the network. 
		Left: Detailed components of the convolutional feature extraction and multi-level feature fusion. 
		The ``ConvUnit(w, h, n)'' represents a convolutional layer of n $w \times h$ kernels, connected by a batch normalization layer and a ReLU layer.
		The ``UpSample(n)'' represents a deconvolution layer of $n \times n$ kernels with stride $n$. 
		Right: The design of multi-task module. ``Scale\&Shift'' is used to stretch and translate the values.}
\end{figure}

By combining outputs of these two tasks, we predict a quadrilateral with score for each point of $\frac{S}{4} \times \frac{S}{4}$ map.
More detailed structure and parameterized configuration of the network is shown in Fig.\hyperref[Fig.network_structure]{4}.

\subsection{Ground Truth and Loss Function}
\label{Sec.3.2}
The full multi-task loss $\mathcal{L}$ can be represented as

\begin{equation}
\mathcal{L} = \mathcal{L}_{cls} + \lambda_{loc} \cdot \mathcal{L}_{loc},
\end{equation}

\noindent where $\mathcal{L}_{cls}$ and $\mathcal{L}_{loc}$ represent loss for classification task and regression task respectively. The balance between two losses is controlled by the hyper-parameter $\lambda_{loc}$.

\begin{figure}
	\label{Fig.gt_input}
	\centering
	\subfigure[]{
		\includegraphics[scale=0.23]{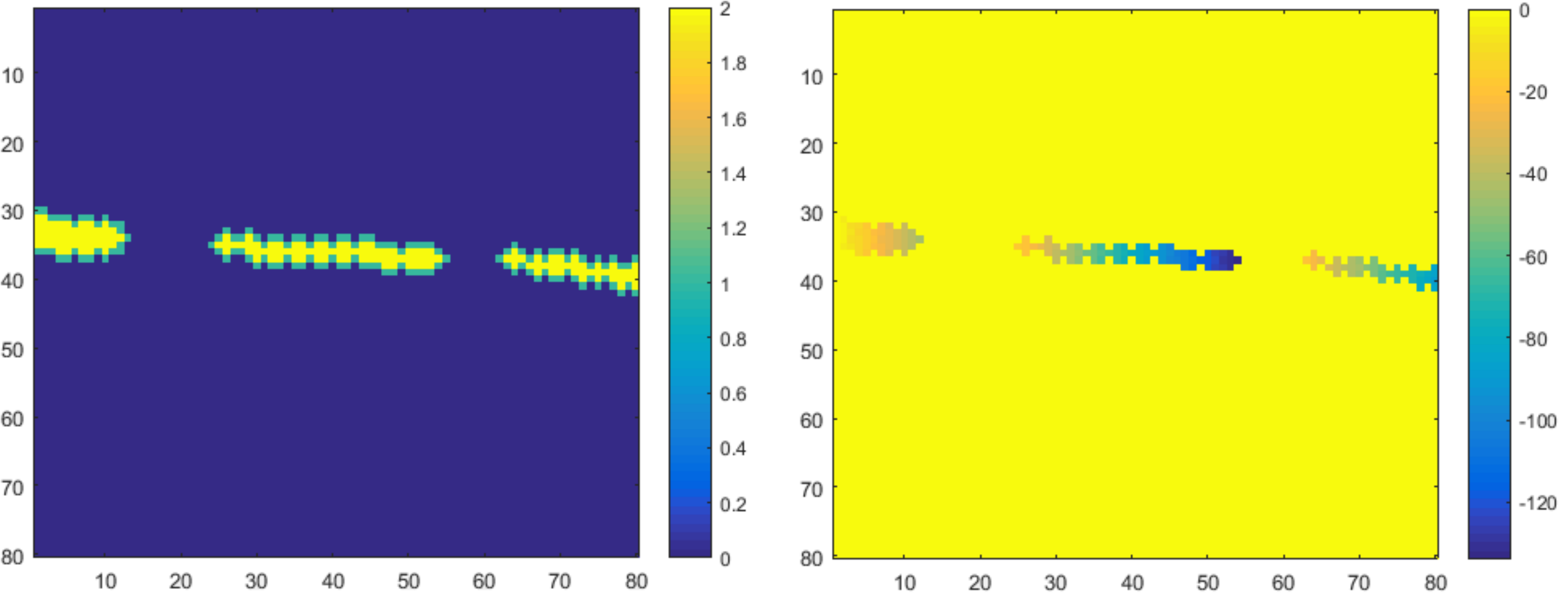}}
	
	\subfigure[]{
		\includegraphics[scale=0.3]{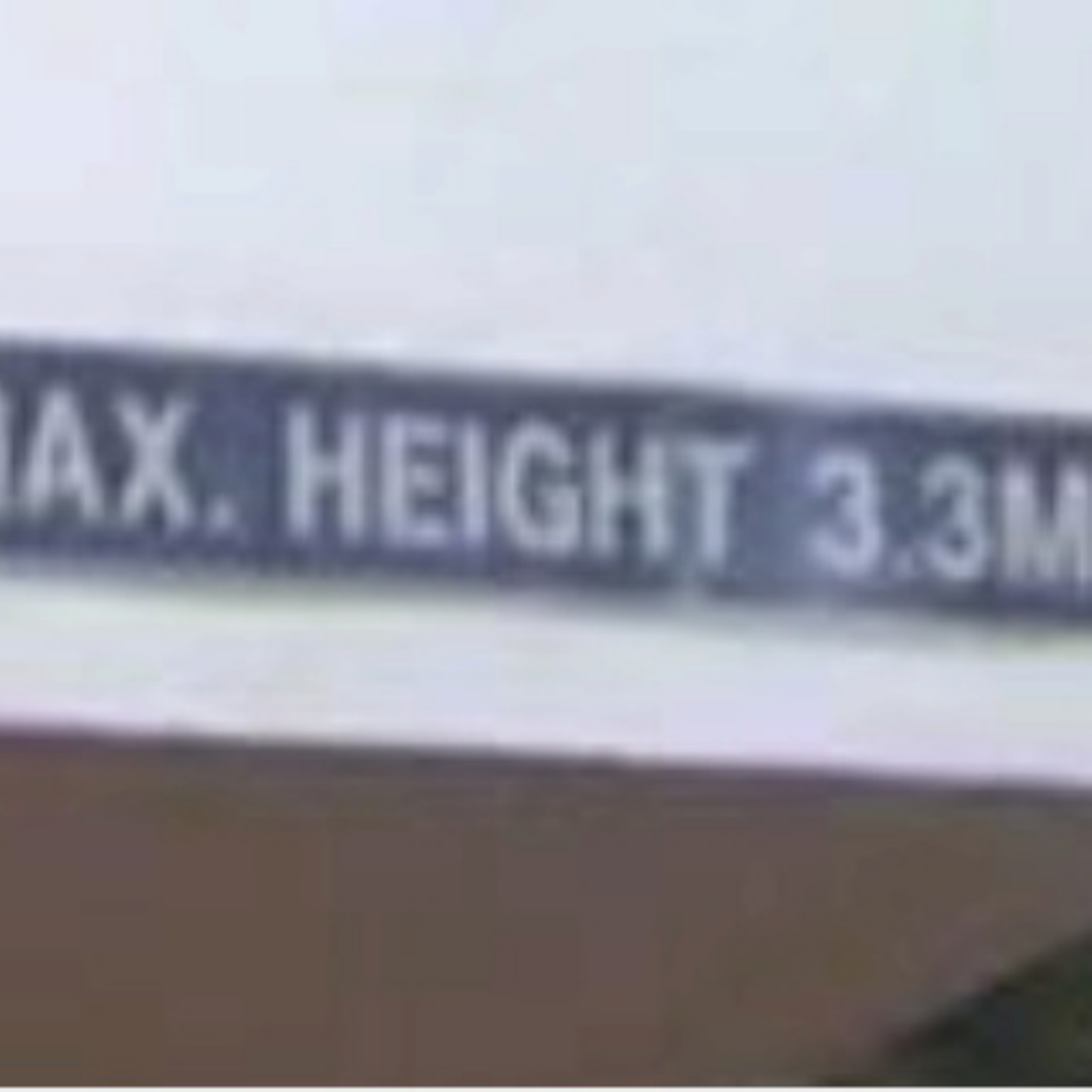}}
	\caption{Visualized ground truths of multi-task. (a) The left map is the ground truth for classification task, where the yellow regions are positive, enclosed by ``NOT CARE'' regions colored in light sea-green. The right map is the ground truth of ``top-left'' channel for regression task. Values grow smaller from left to right within a word region as pixels are farther from the top left corner. (b) The corresponding input image of the ground truths.}
\end{figure}

\noindent \textbf{Classification task.}
Although the ground truth for classification task can be deemed as a down-sampled segmentation between text and non-text, unlike the implementation in \cite{fcn-text}, we do not take all pixels within text region as positive, instead, we only regard pixels around the text center line within distance $r$ as positive and enclose positive region with an ``NOT CARE'' boundary as transition from positive to negative (shown in Fig.\hyperref[Fig.gt_input]{5}). The parameter $r$ is proportional to the short side of text boundaries and its value is 0.2.

Furthermore, text is taken as a positive sample only when its short side length ranges in $\left[32 \times 2^{-1}, 32 \times 2^{1} \right]$. If the short side length falls in $\left[32 \times 2^{-1.5},  32 \times 2^{-1}\right) \cup \left(32 \times 2^{1},  32 \times 2^{1.5}\right]$, we take the text as ``NOT CARE'', otherwise negative. ``NOT CARE'' regions do not contribute to the training objective. Ground truths designed in this way reduce the confusion between text and non-text, which is beneficial for discriminative feature learning.

The loss function $\mathcal{L}_{cls}$ chosen for classification task is the hinge loss. Denote the ground truth for a given pixel as $y^{\ast}_{i} \in \left\{0, 1\right\}$ and predicted value as $\hat{y_{i}}$, $\mathcal{L}_{cls}$ is formulated as

\begin{equation}
	\mathcal{L}_{cls} = \frac{1}{S^{2}} \sum_{i \in \mathcal{L}_{cls}} \text{max}\left(0, \text{sign}\left(0.5 - y^{\ast}_{i}\right) \cdot \left(\hat{y_{i}} - y^{\ast}_{i} \right) \right)^{2}
\end{equation}


Besides this, we also adopt the class balancing and hard negative sample mining as introduced in \cite{densebox} for better performance and faster loss convergence. Hence during training, the predicted values for ``NOT CARE'' region and easily classified negative area are forced to zero, the same as the ground truth.

\noindent \textbf{Regression task.}
Considering that the ground truth values of regression task vary within a wide range, we use a $Scale \& Shift$ module (shown in Fig.\hyperref[Fig.network_structure]{4}) for fast convergence. $Scale \& Shift$ takes the value $z$ from a sigmoid neuron as input and stretch $z$ into $ \hat{z}$ by

\begin{equation}
\hat{z} = 800 \cdot z - 400, \ z \in \left(0, 1\right)
\end{equation}

Here we assume that the maximum positive text size is less than 400. We also use a sigmoid neuron to normalize the values before $Scale \& Shift$ for steady convergence.

According to \cite{fast-rcnn}, the loss function $\mathcal{L}_{loc}$ used in regression task is defined as follows. Denote the ground truth for a given pixel as $z^{\ast}_{i}$ and predicted value as $\hat{z_{i}}$, $\mathcal{L}_{loc}$ is formulated as 

\begin{equation}
\mathcal{L}_{loc} = \sum_{i \in \mathcal{L}_{loc}}^{} \left[y^{\ast}_i > 0\right] \cdot \text{smooth}_{L_1} \left(z^{\ast}_{i} - \hat{z_{i}}\right) ,
\end{equation}


\begin{equation}
\text{smooth}_{L_1}\left(x\right) = 
\left\{\begin{matrix}
0.5x^2 & \text{if} \left | x \right |  < 1 , \\ 
\left |x\right | - 0.5 & \text{otherwise} .
\end{matrix}\right.
\end{equation}

We choose smooth $L_1$ loss here because it is less sensitive to outliers compared with $L_2$ loss. During training stage, smooth $L_1$ loss need less careful tuning of learning rate and decreases steadily.

\subsection{Recalled Non-Maximum Suppression}
\label{Sec.3.3}
After getting the outputs produced by multi-task learning, each point of the output map is related with a scored quadrilateral.
To filter the non-text region, we only preserve points with high score in classification task. However, there will be still densely overlapped quadrilaterals for a word or text line. To reduce the redundant results we propose a post-processing method called Recalled Non-Maximum Suppression. 

The Recalled NMS is a trade-off solution for two problems: (i) when texts are close, quadrilaterals between two words are often retained because of the difficulty in classifying pixels within word space, (ii) if we solve problem (i) by simply retaining quadrilaterals with higher score, text region with relative lower confidence will be discarded and the overall recall will be sacrificed a lot. The Recalled NMS could both remove quadrilaterals within text spaces and maintain the text region with low confidence.

\begin{figure}[H]
	\label{Fig.recalled_nms}
	\centering
	\includegraphics[scale=0.1]{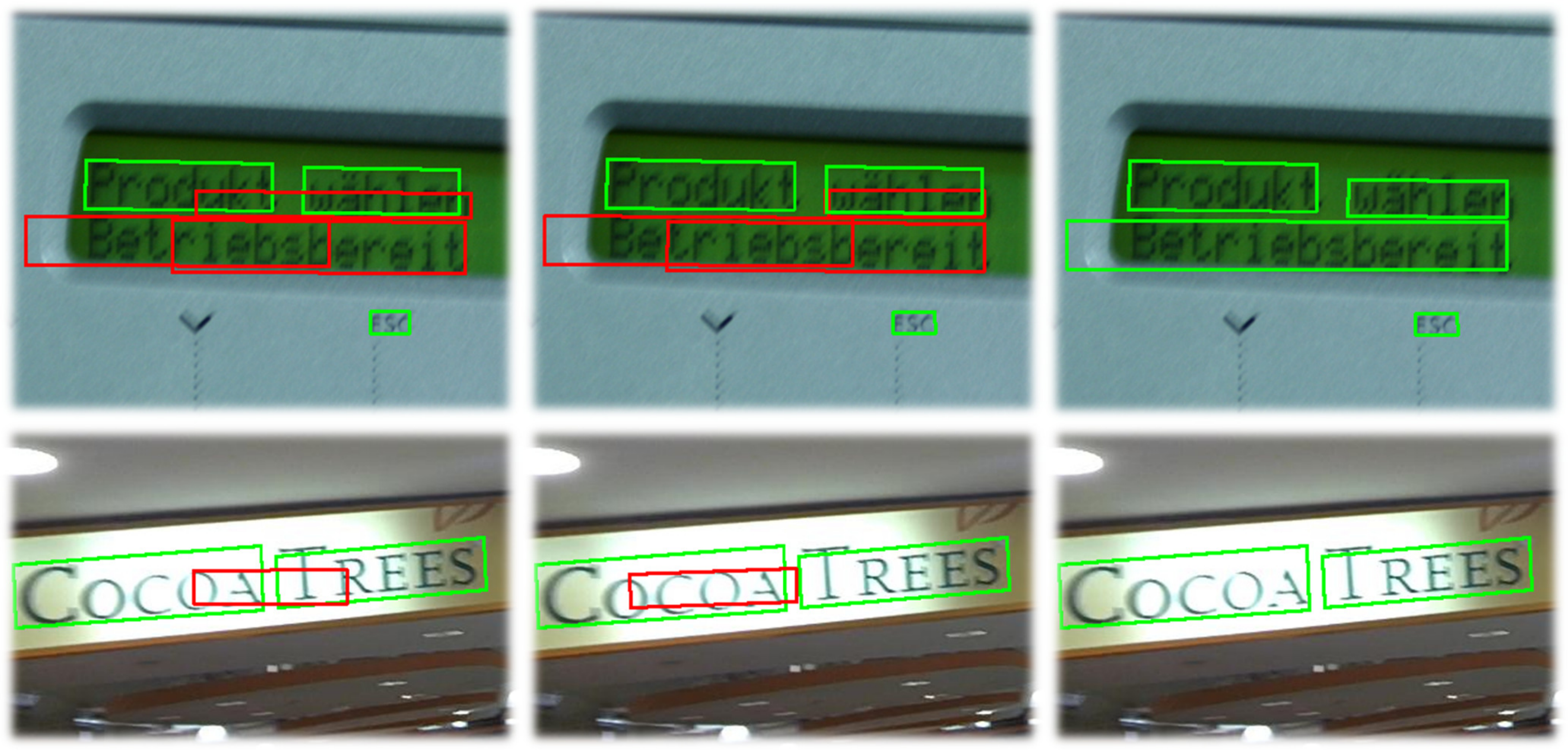}
	\caption{Three steps in Recalled NMS. Left: results of traditional NMS (quadrilaterals in red are false detection). Middle: recalled high score quadrilaterals. Right: merging results by closeness.}
\end{figure}

The Recalled NMS has three steps as shown in Fig.\hyperref[Fig.recalled_nms]{6}.
\begin{itemize}
	\setlength{\itemsep}{0pt}
	\setlength{\parsep}{0pt}
	\setlength{\parskip}{0pt}
	\item First, we get suppressed quadrilaterals $\mathcal{B}_{sup}$ from densely overlapped quadrilaterals $\mathcal{B}$ by traditional NMS.
	\item Second, each quadrilateral in $\mathcal{B}_{sup}$ is switched to the one with highest score in $\mathcal{B}$ beyond a given overlap. After this step, quadrilaterals within word space are changed to those of higher score and low confidence text region are preserved as well.
	\item Third, after the second step we may get dense overlapped quadrilaterals again, and instead of suppression, we merge quadrilaterals in $\mathcal{B}_{sup}$ which are close to each other.
	
\end{itemize}

\subsection{Network Implementation}
\label{Sec.3.4}
The training samples of $320 \times 320$ are cropped from scaled images rotated randomly by $0, \ {\pi}/{2}, \ \pi, \ \text{or} \ {3\pi}/{2}$. 
The task balance index $\lambda_{loc}$ is raised from 0.01 to 0.5 after the classification task gets well trained. The network should learn what the text is first and then learn to localize the text.
In testing, we adopt a multi-scale sliding window strategy in which window size is $320 \times 320$, sliding stride is $160$ and multi-scale set is $\left\{2^{-5}, 2^{-4}, \cdots, 2^{1}\right\}$. 
Pixels on $\mathcal{M}_{cls}$ are deemed as text if their values are higher than $0.7$. 
In post processing, the only parameter, overlap ratio, in Recalled NMS is 0.5.

\section{Experiments}
\label{Sec.4}
We evaluate our method on three benchmarks: ICDAR2015 Incidental Scene Text, MSRA-TD500 and ICDAR2013. The first two datasets have multi-oriented texts and the third one has mostly horizontal texts. For fair comparison we also list recent state-of-the-art methods on these benchmarks.

\subsection{Benchmark Description}
\label{Sec.4.1}

\noindent \textbf{ICDAR2015 Incidental Scene Text.} This dataset is recently published for ICDAR2015 Robust Reading Competition. It contains 1000 training images and 500 test images. Different from previous scene text datasets where texts are well captured in high resolution, this dataset contains texts with various scales, resolution, blurring, orientations and viewpoint. 
The annotation of bounding box (actually quadrilateral) also differs greatly from previous ones which has 8 coordinates of four corners in a clock-wise manner. 
In evaluation stage, word-level predictions are required.

\noindent \textbf{MSRA-TD500.} This dataset contains 300 training images and 200 test images, where there are many multi-oriented text lines. Texts in this dataset are stably captured with high resolution and are bi-lingual of both English and Chinese.

The annotations of MSRA-TD500 are at line level which casts great influence on optimizing regression task.
Lacking of line level annotation and sufficient bi-lingual training data, we did not use the training set and instead, we utilize the generalization of our model trained on English word-level data.

\noindent \textbf{ICDAR2013 Focused Scene Text.} This dataset lays more emphasis on horizontal scene texts. It contains 229 training images and 233 test images which are well captured and clear.	The evaluation protocol is introduced in \cite{karatzas2013icdar}.

\subsection{Implementation Details}
\label{Sec.4.2}
The network is optimized by stochastic gradient descent (SGD) with back-propagation and the max iteration is $2 \times 10^{5}$. We adopt the ``multistep'' strategy in Caffe \cite{caffe} to adjust learning rate. For the first $3 \times 10^{4}$ iterations the learning rate is fixed to be $10^{\text{-}2}$ and after that it is reduced to $10^{\text{-}3}$ until the $10^5$th iteration. For the rest $10^5$ iterations, the learning rate keeps $10^{\text{-}4}$. Apart from adjusting learning rate, the hard sample ratio mentioned in Sec.\hyperref[Sec.3.2]{3.2} is increased from 0.2 to 0.7 at the $3 \times 10^4$th iteration. Weight decay is $4 \times 10^{\text{-}4}$ and momentum is 0.9. All layers except in regression task are initialized by ``xavier'' \cite{glorot2010understanding} and the rest layers are initialized to a constant value zero for stable convergence.

The model is optimized on training datasets from ICDAR2013 and ICDAR2015, as well as 200 negative images (scene images without text) collected from the Internet. The whole experiments are conducted on Caffe and run on a workstation with 2.9GHz 12-core CPU, 256G RAM, GTX Titan X and Ubuntu 64-bit OS.

\subsection{Experimental Results}
\label{Sec.4.3}

\noindent \textbf{ICDAR2015 Incidental Scene Text.} 
The results shown in Tab.\hyperref[Tab.1]{1} indicates that the proposed method outperforms previous approaches by a large margin in both precision and recall.
To demonstrate the effectiveness of Recalled NMS, we also list the result adopting traditional NMS as the post processing. From Tab.\hyperref[Tab.1]{1} we can see the Recalled NMS give a higher precision mainly because of filtering quadrilaterals between text lines.

\begin{figure*}
	\label{Fig.icdar15_result}
	\centering
		\includegraphics[scale=0.09]{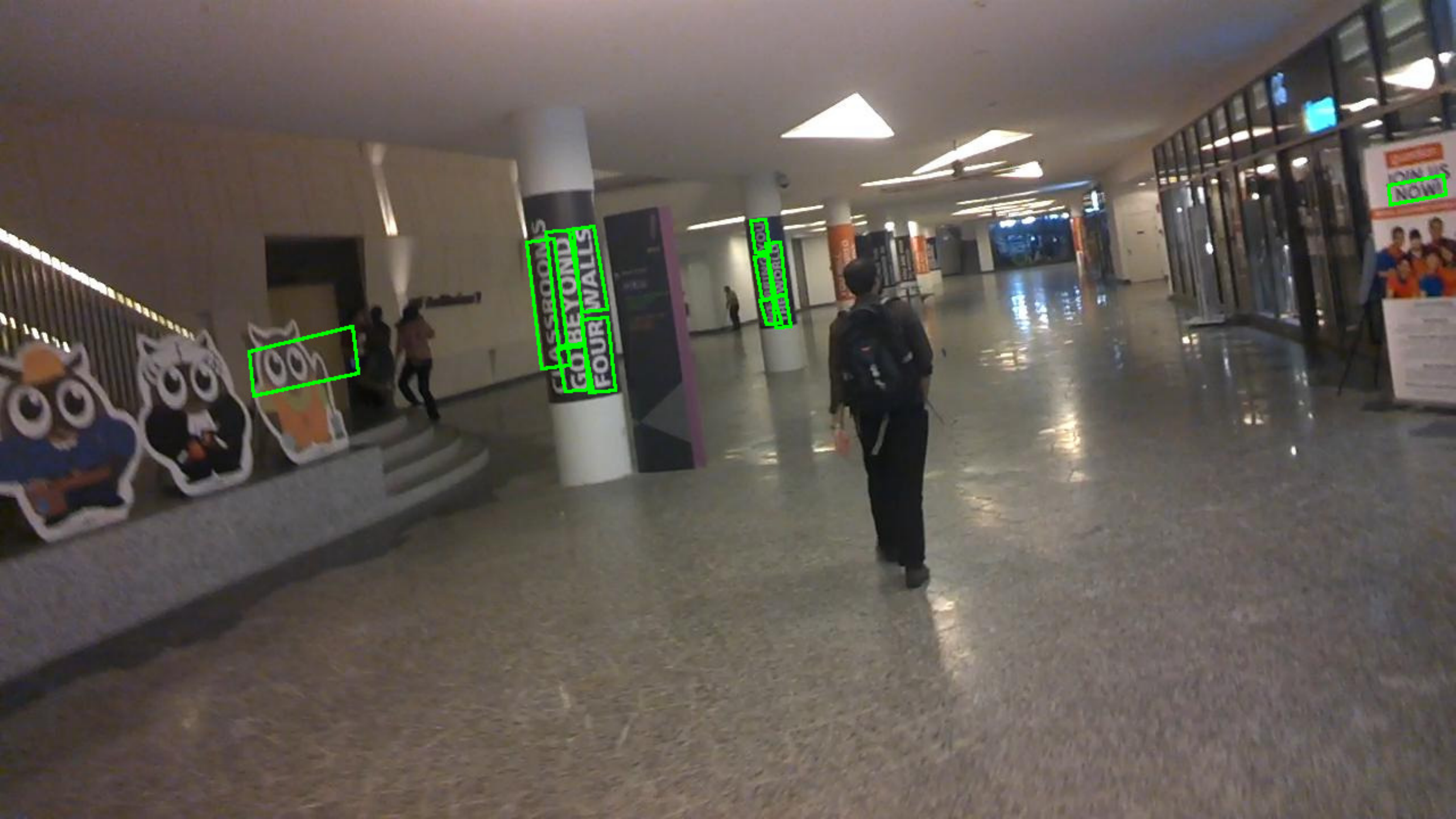}
		\includegraphics[scale=0.09]{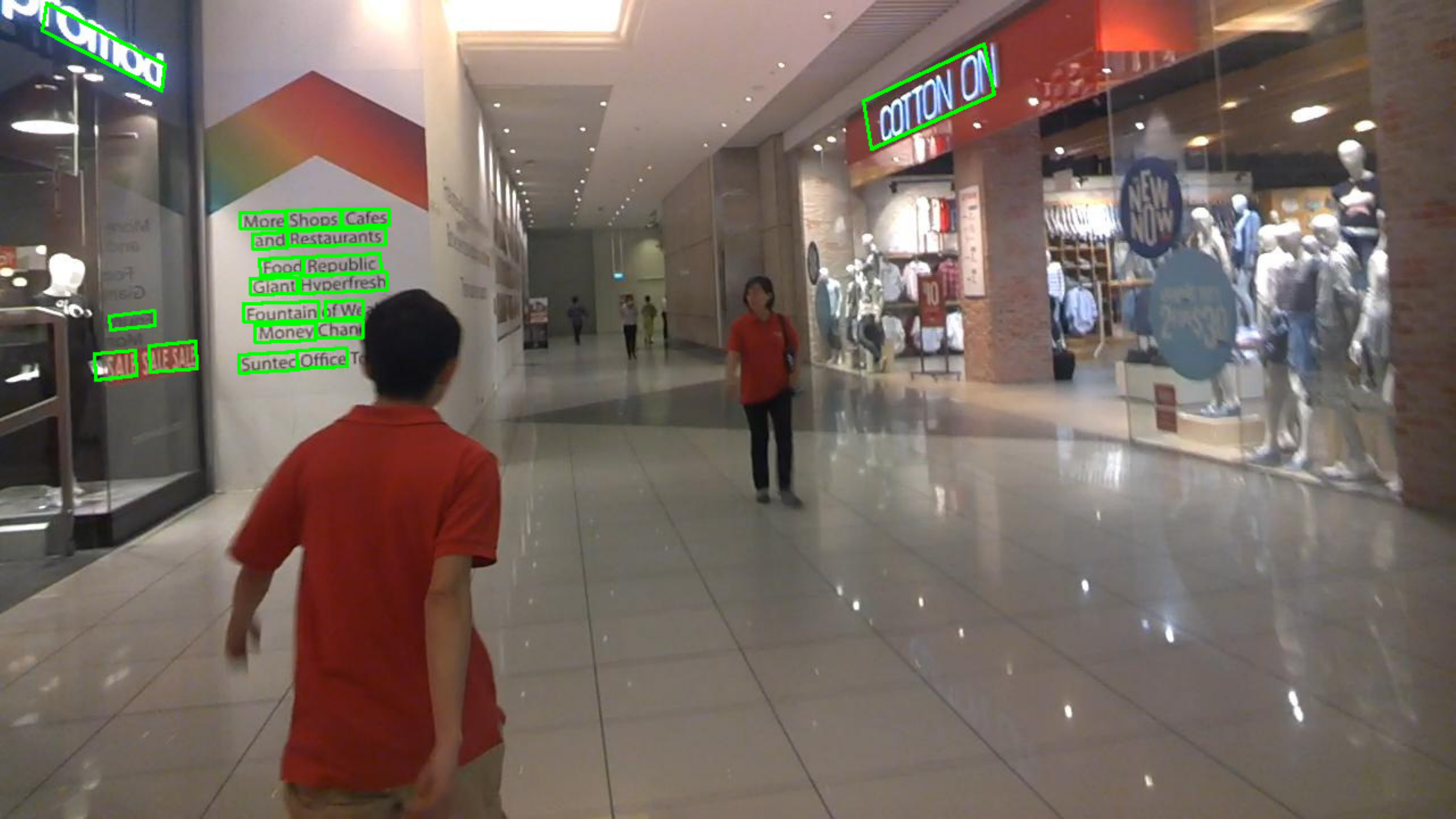}
		\includegraphics[scale=0.09]{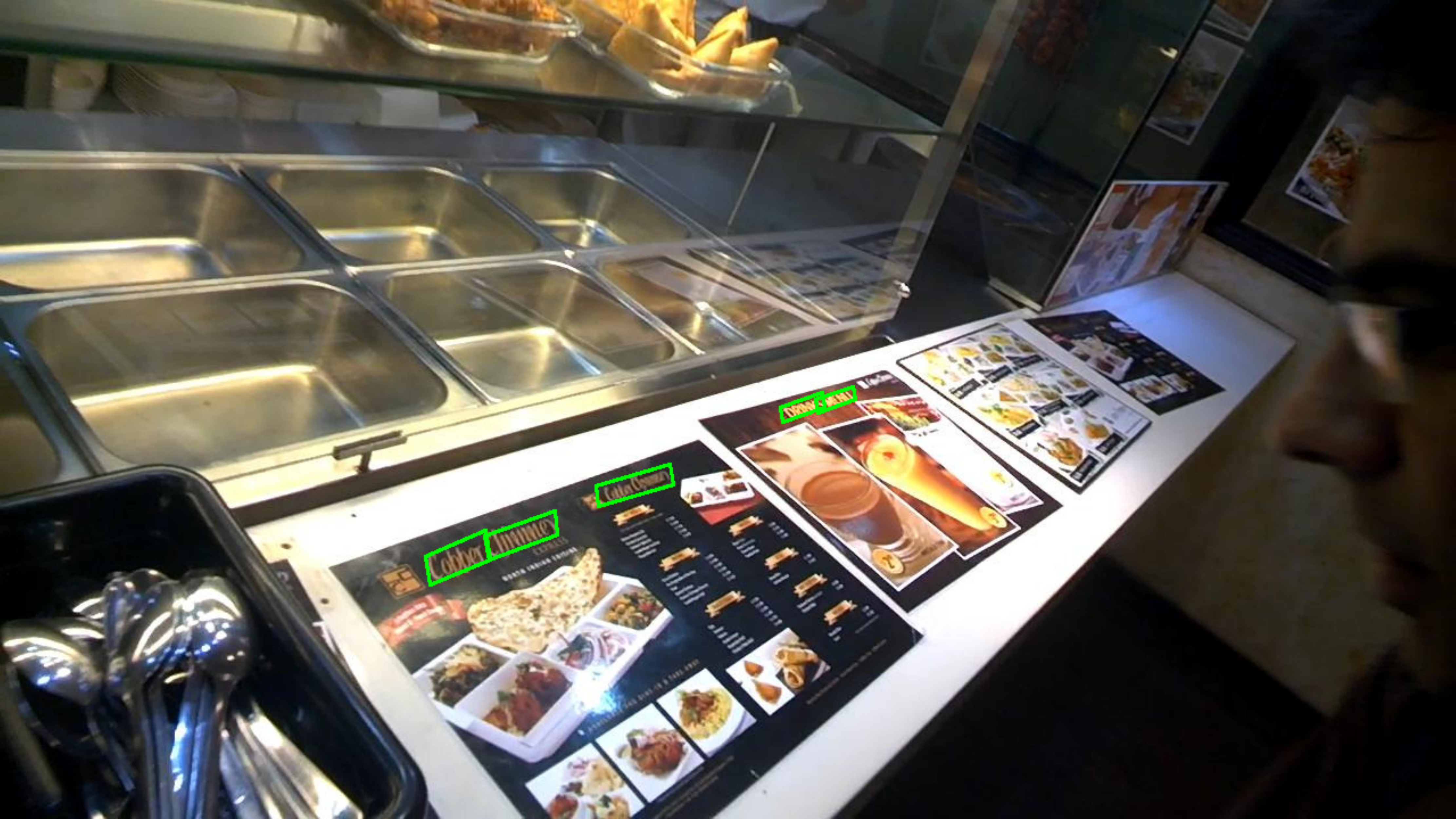}
		\includegraphics[scale=0.09]{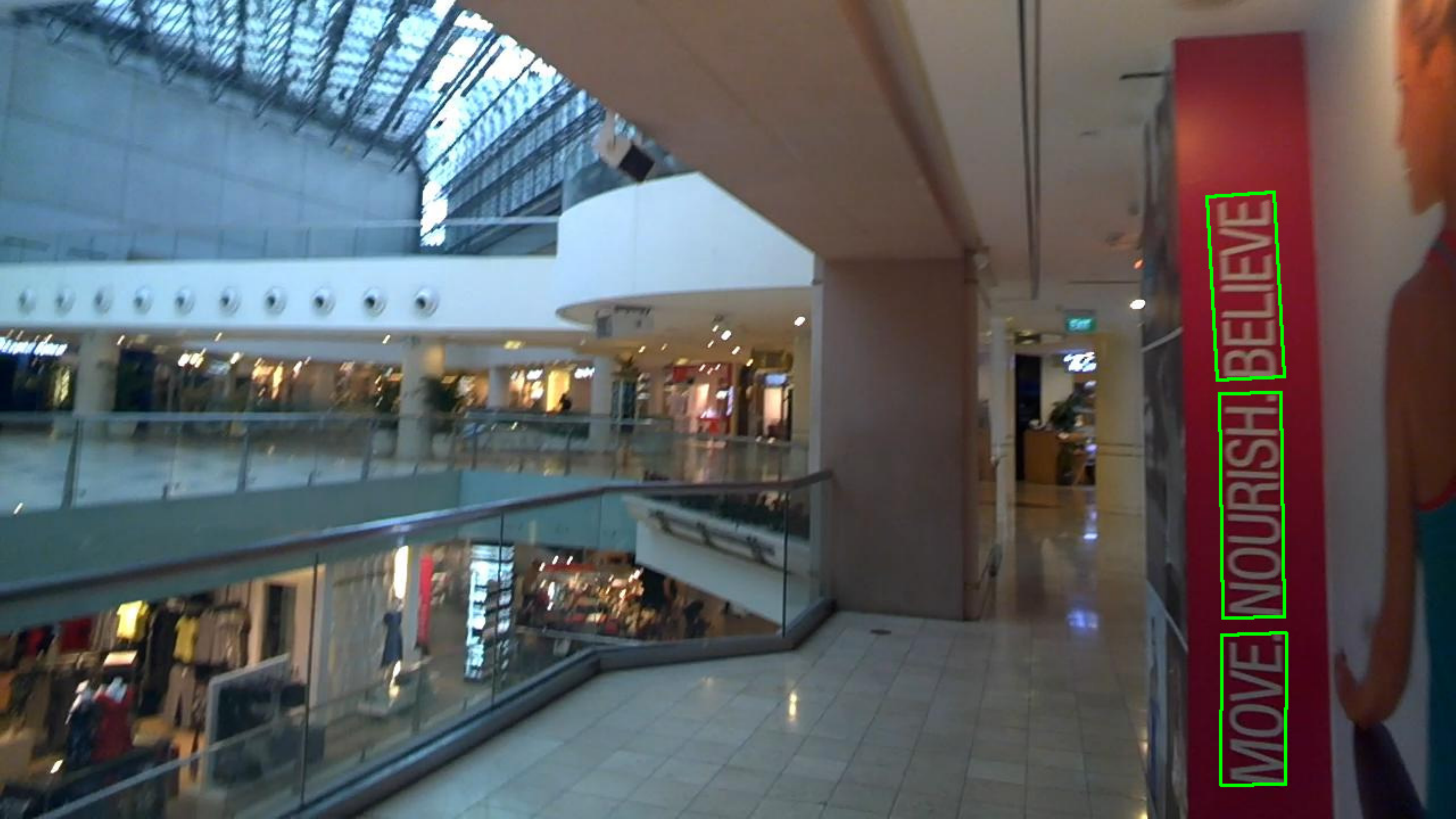}
		\includegraphics[scale=0.09]{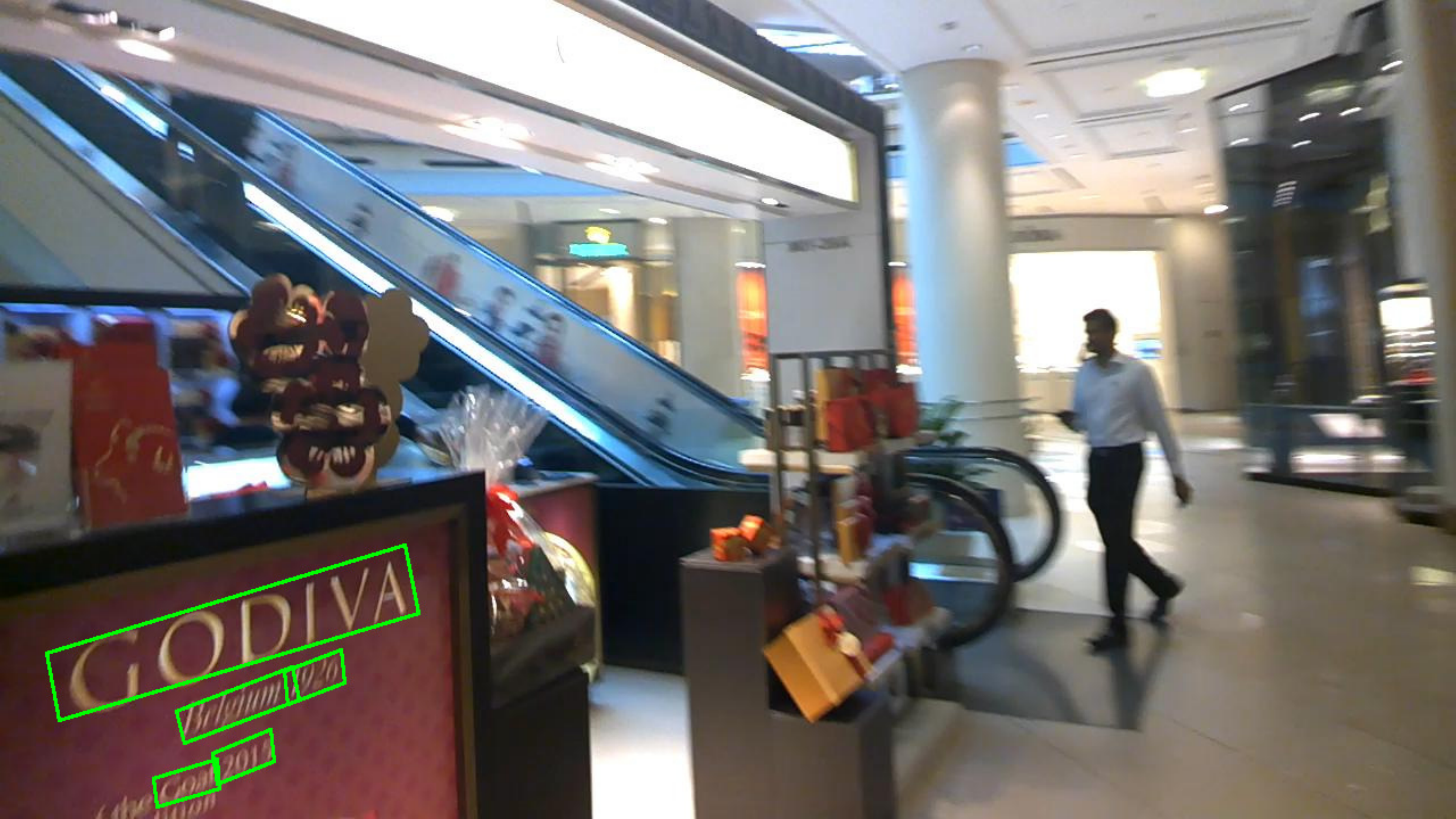}
		\includegraphics[scale=0.09]{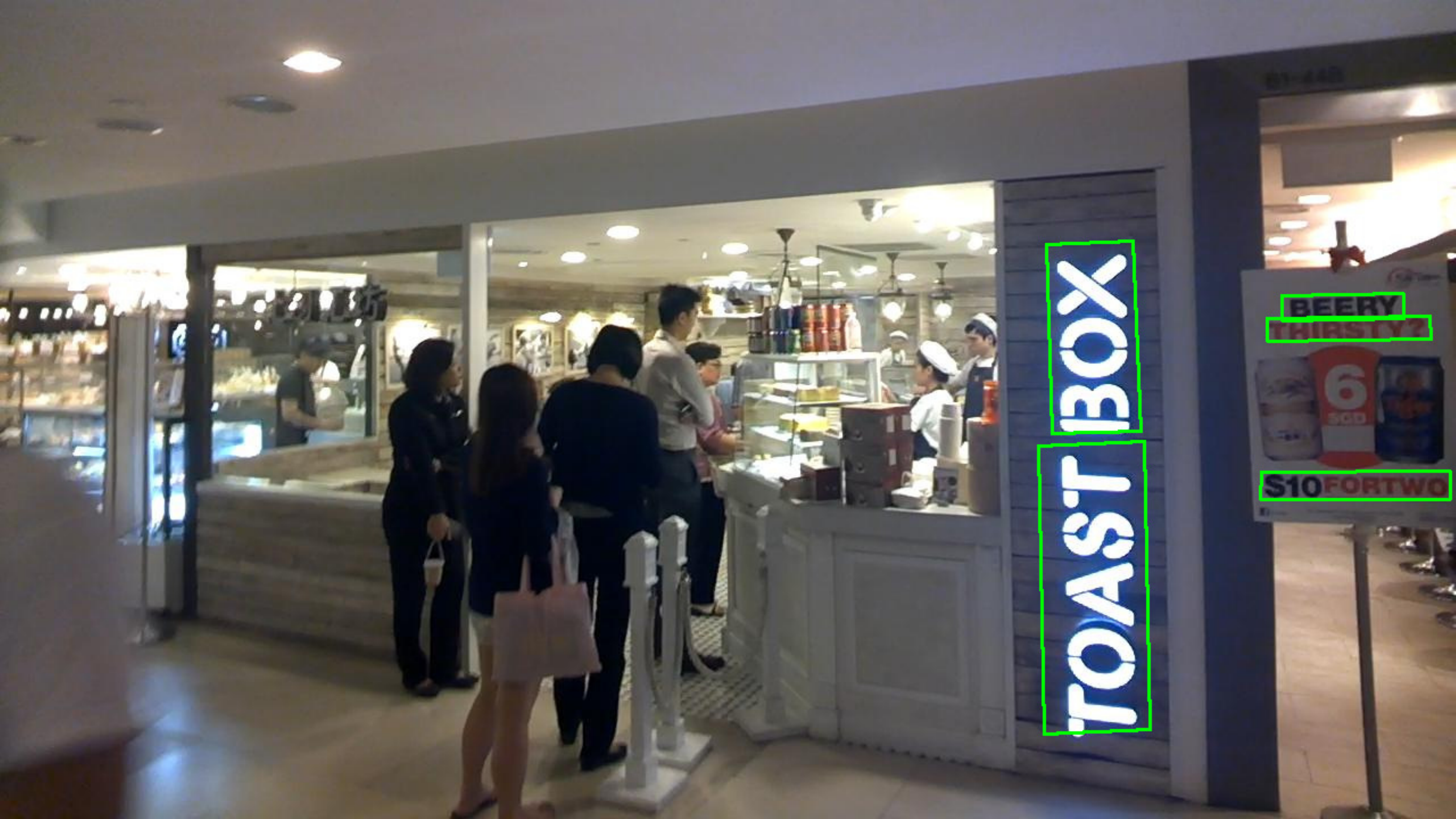}
		\includegraphics[scale=0.09]{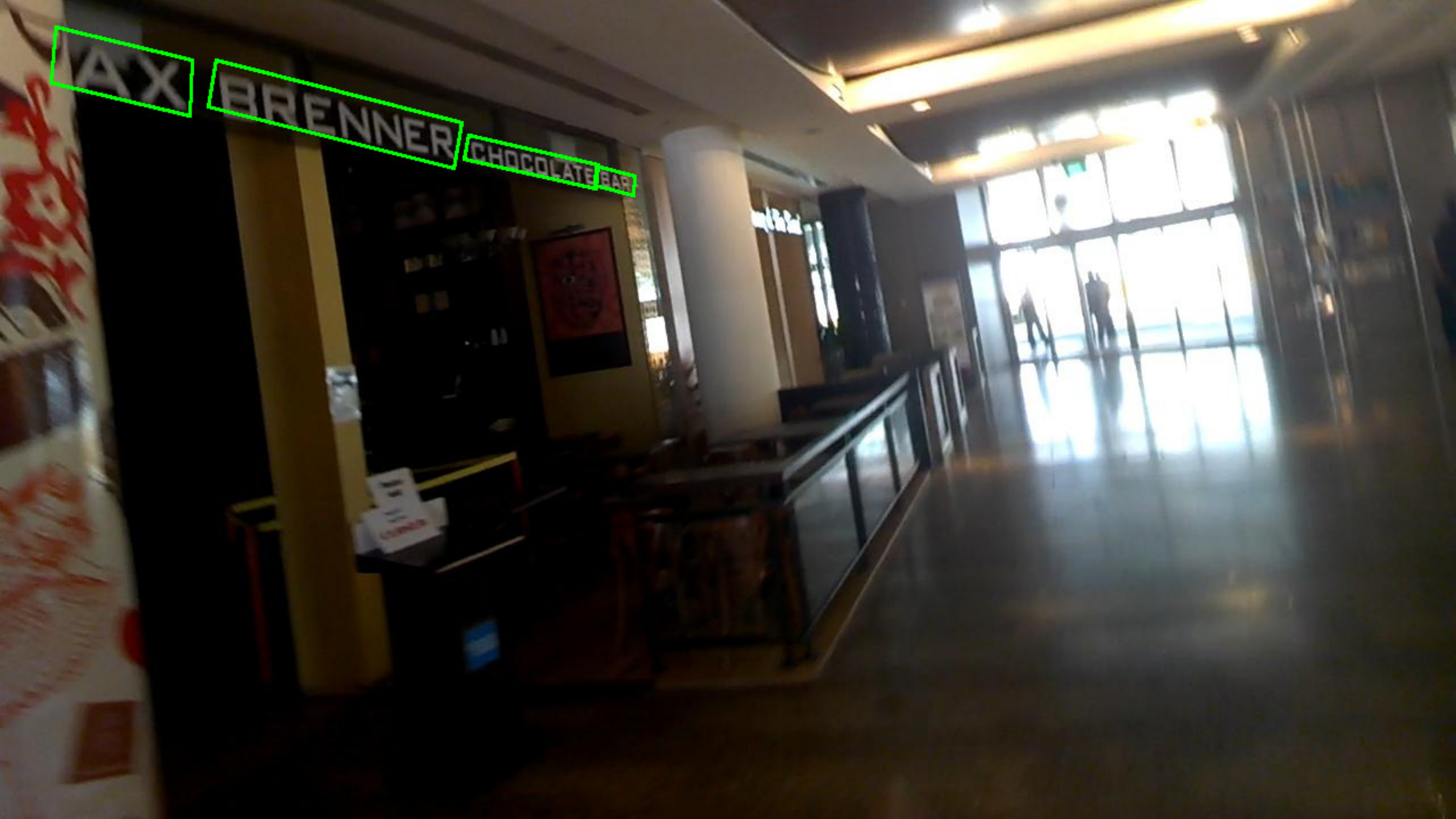}
		\includegraphics[scale=0.09]{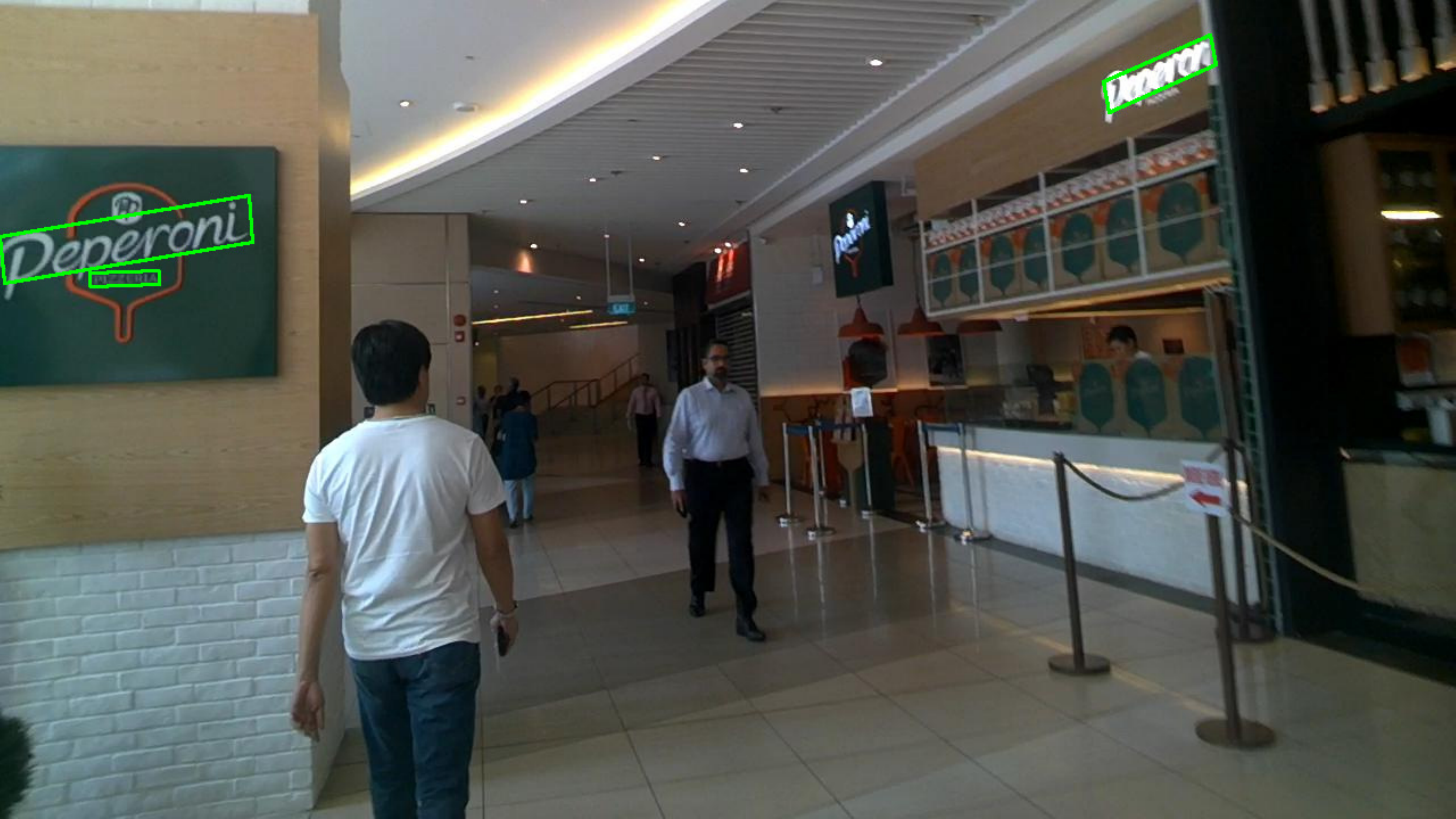}
	
	\caption{Detection examples of our model on ICDAR2015 Incidental Scene Text benchmark.}
\end{figure*}

Note that the method in \cite{dmpn} which ranks second is indirect regression based multi-oriented text detection and it also treats text detection as object detection. The large margin between our method and this method demonstrates our analysis on the deficiency of indirect regression and superiority of direct regression for multi-oriented text detection.
Some examples of our detection results are shown in Fig.\hyperref[Fig.icdar15_result]{7}.

\begin{table}
	\label{Tab.1}
	\small
	\renewcommand\arraystretch{1.2}
	\centering
	\caption{Comparison of methods on ICDAR2015 Incidental Scene Text dataset. R-NMS is short for Recalled NMS and T-NMS is short for traditional NMS.}
	\begin{tabular}{|c|c|c|c|}
		\hline
		Algorithm & Precision & Recall & F-measure \\
		\hline
		\hline
		{Proposed (R-NMS)} & \textbf{0.82} & \textbf{0.80} & \textbf{0.81} \\
		\hline
		{Proposed (T-NMS)} & {0.81} & {0.80} & {0.80} \\
		\hline
		Liu \emph{et al.} \cite{dmpn} & {0.73} & {0.68} & {0.71} \\
		\hline
		Tian \emph{et al.} \cite{ctpn} & 0.74 & 0.52 & 0.61 \\
		\hline
		Zhang \emph{et al.} \cite{fcn-text} & 0.71 & 0.43 & 0.54 \\
		\hline
		StradVision2 \cite{karatzas2015icdar} & 0.77 & 0.37 & 0.50 \\
		\hline
		StradVision1 \cite{karatzas2015icdar} & 0.53 & 0.46 & 0.50 \\
		\hline
		NJU-Text \cite{karatzas2015icdar} & 0.70 & 0.36 & 0.47 \\
		\hline
		AJOU \cite{karatzas2015icdar} & 0.47 & 0.47 & 0.47 \\
		\hline
		HUST\_MCLAB  \cite{karatzas2015icdar} & 0.44 & 0.38 & 0.41 \\
		\hline
	\end{tabular}
\end{table}

\begin{figure}
	\label{Fig.td500_result}
	\centering	
	\subfigure[] {
		\label{Fig.sub.td500_result_1}
		\includegraphics[width=4cm, height=3cm]{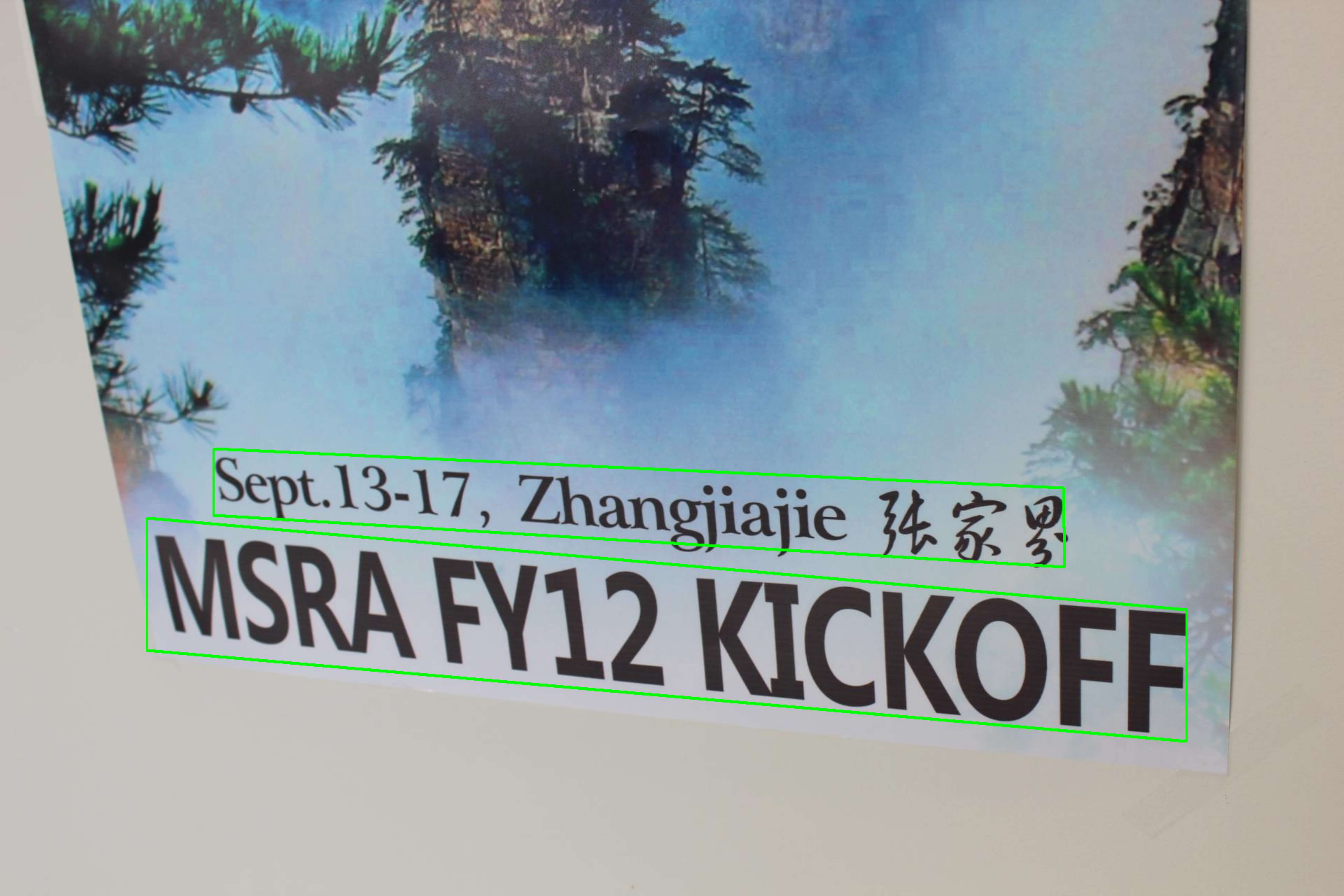}}
	\subfigure[] {
		\label{Fig.sub.td500_result_3}
		\includegraphics[width=4cm, height=3cm]{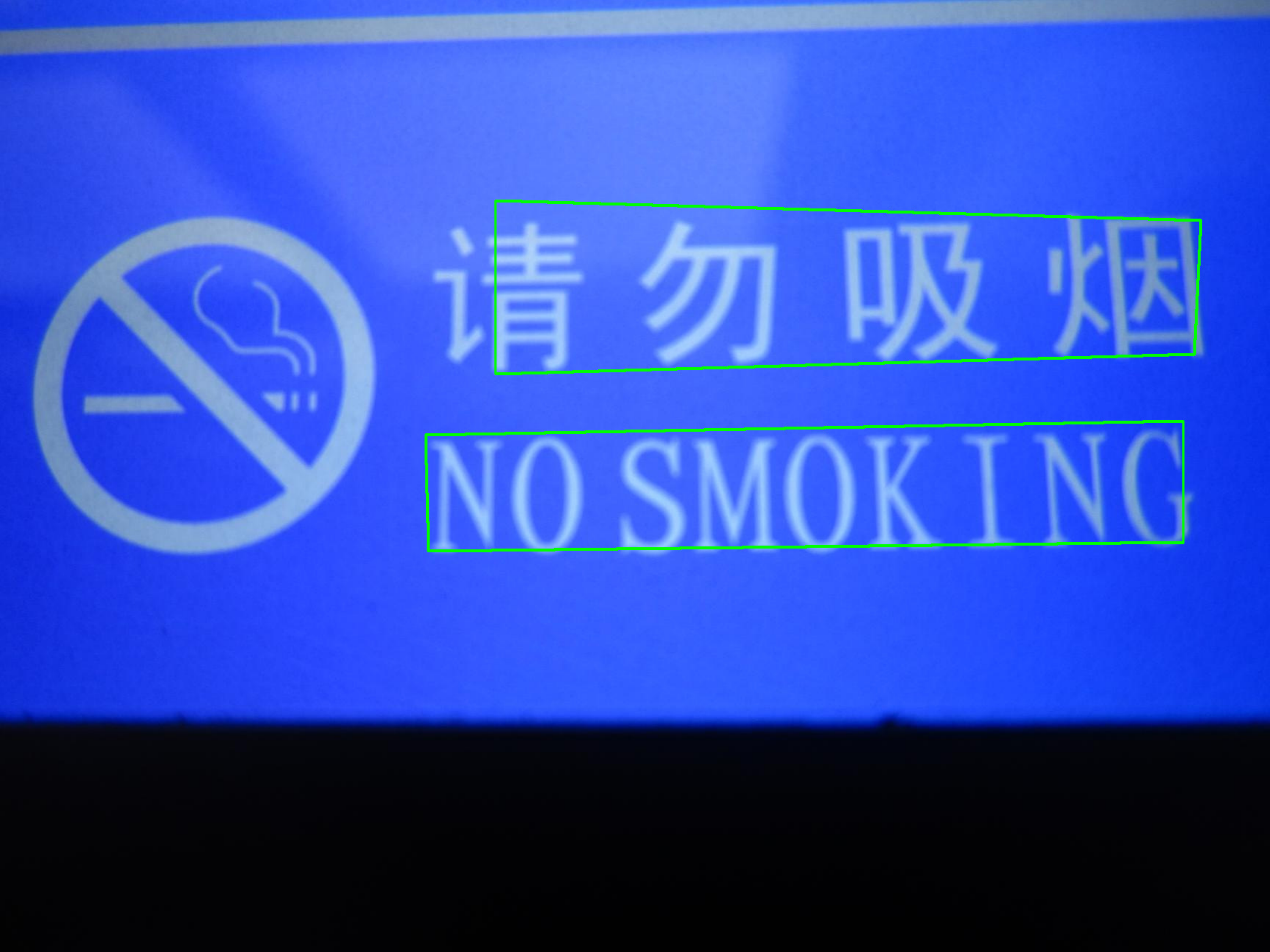}}
	\subfigure[] {
		\label{Fig.sub.td500_result_7}
		\includegraphics[width=4cm, height=3cm]{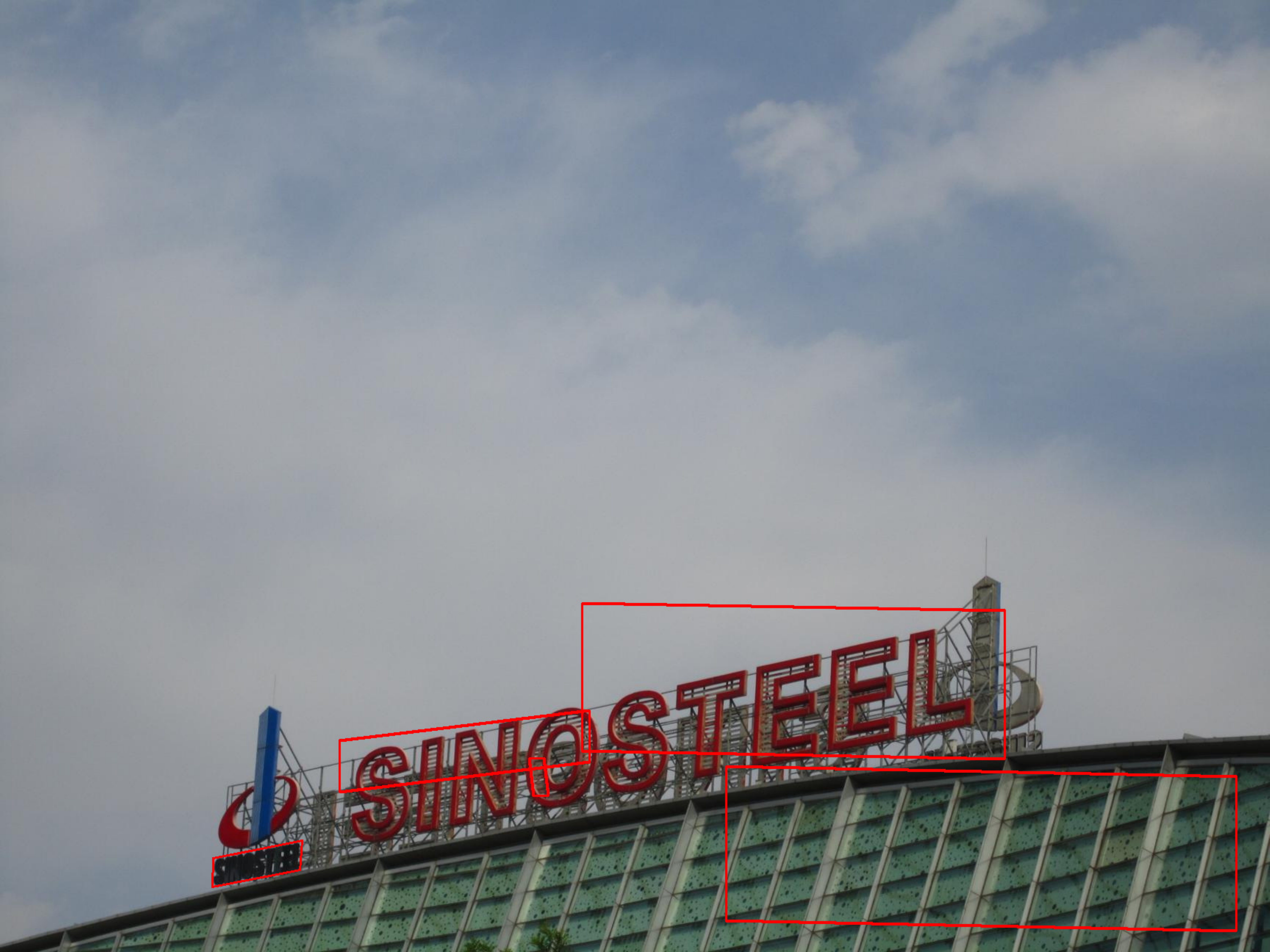}}
	\subfigure[] {
		\label{Fig.sub.td500_result_8}
		\includegraphics[width=4cm, height=3cm]{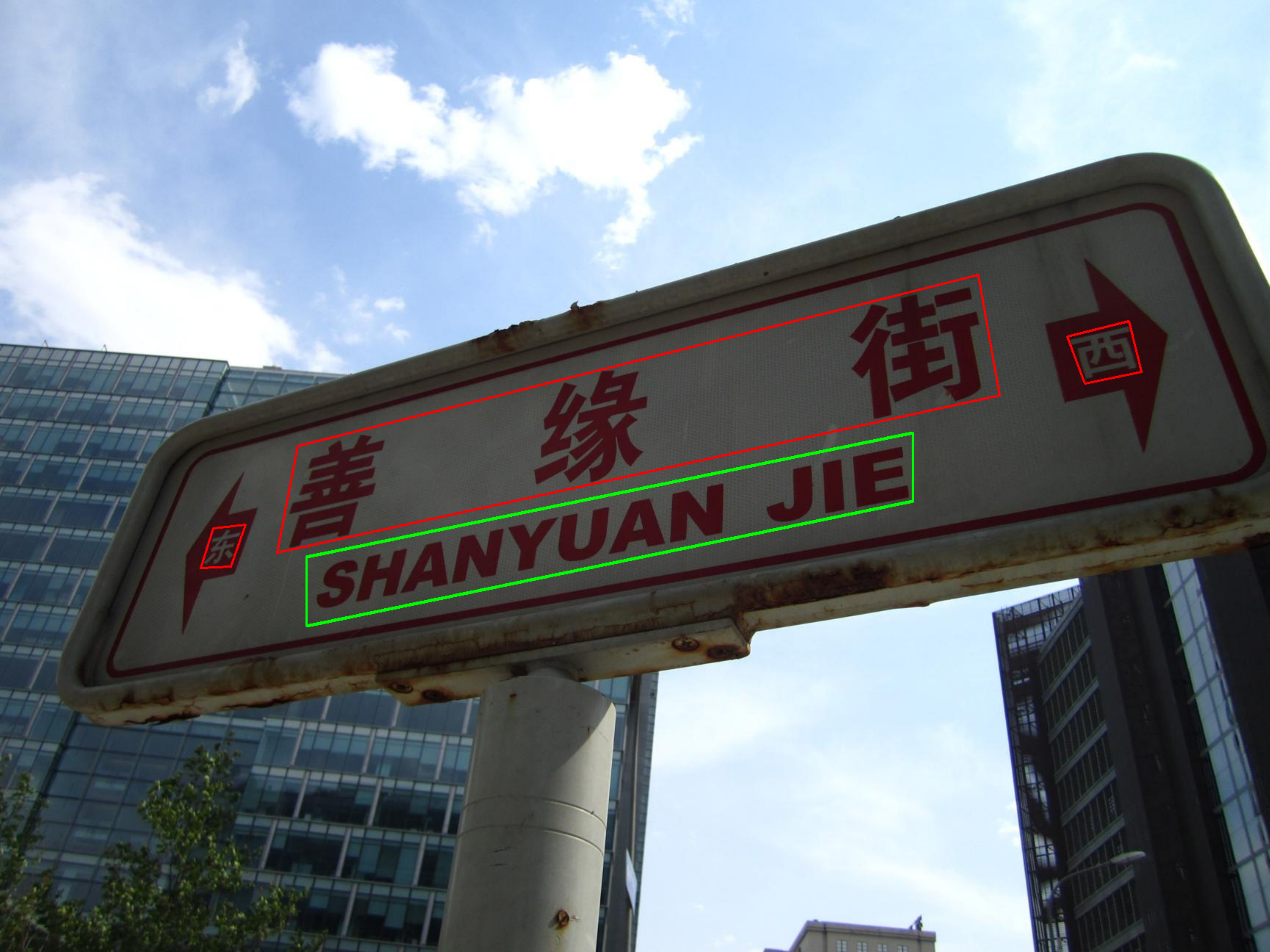}}
	
	\caption{Detection examples of our model on MSRA-TD500. (a)-(b) Chinese text can also be detected due to the model generalization. (c)-(d) Some failure cases for complicated background or wide character space. False and miss detected texts are enclosed by red lines.}
\end{figure}

\noindent \textbf{MSRA-TD500.} The results of our method on this dataset are shown in Tab.\hyperref[Tab.2]{2}, with comparisons to other representative results of state-of-the art methods.
It is shown that our method could reach the state-of-the-art performance. It should be noted that we did not adopt the provided training set or any other Chinese text data.
Since our method could only detect text in word level, we implement line grouping method based on heuristic rules in post processing.
Our model shows strong compatibility for both English and Chinese, however, we still fail to detect Chinese text lines that have wide character spaces or complex background. 
Part of our detection results are shown in Fig.\hyperref[Fig.td500_result]{8}.

\begin{table}
	\label{Tab.2}
	\small
	\renewcommand\arraystretch{1.2}
	\centering
	\caption{Comparison of methods on MSRA-TD500 dataset.}
	\begin{tabular}{|c|c|c|c|}
		\hline
		Algorithm & Precision & Recall & F-measure \\
		\hline
		\hline
		Proposed & 0.77 & \textbf{0.70} & \textbf{0.74} \\
		\hline
		Zhang \emph{et al.} \cite{fcn-text} & \textbf{0.83} & 0.67 & 0.74 \\
		\hline
		Yin \emph{et al.} \cite{yin2015multi} & 0.81 & 0.63 & 0.71 \\ 
		\hline
		Kang \emph{et al.} \cite{kang2014orientation} & 0.71 & 0.62 & 0.66 \\ 
		\hline
		Yao \emph{et al.} \cite{yao2012detecting} & 0.63 & 0.63 & 0.60 \\ 
		\hline
	\end{tabular}
\end{table}

\begin{figure}
	\label{Fig.icdar13_result}
	\centering
	\subfigure[] {
		\label{Fig.sub.icdar13_result_1}
		\includegraphics[width=2.5cm, height=3cm]{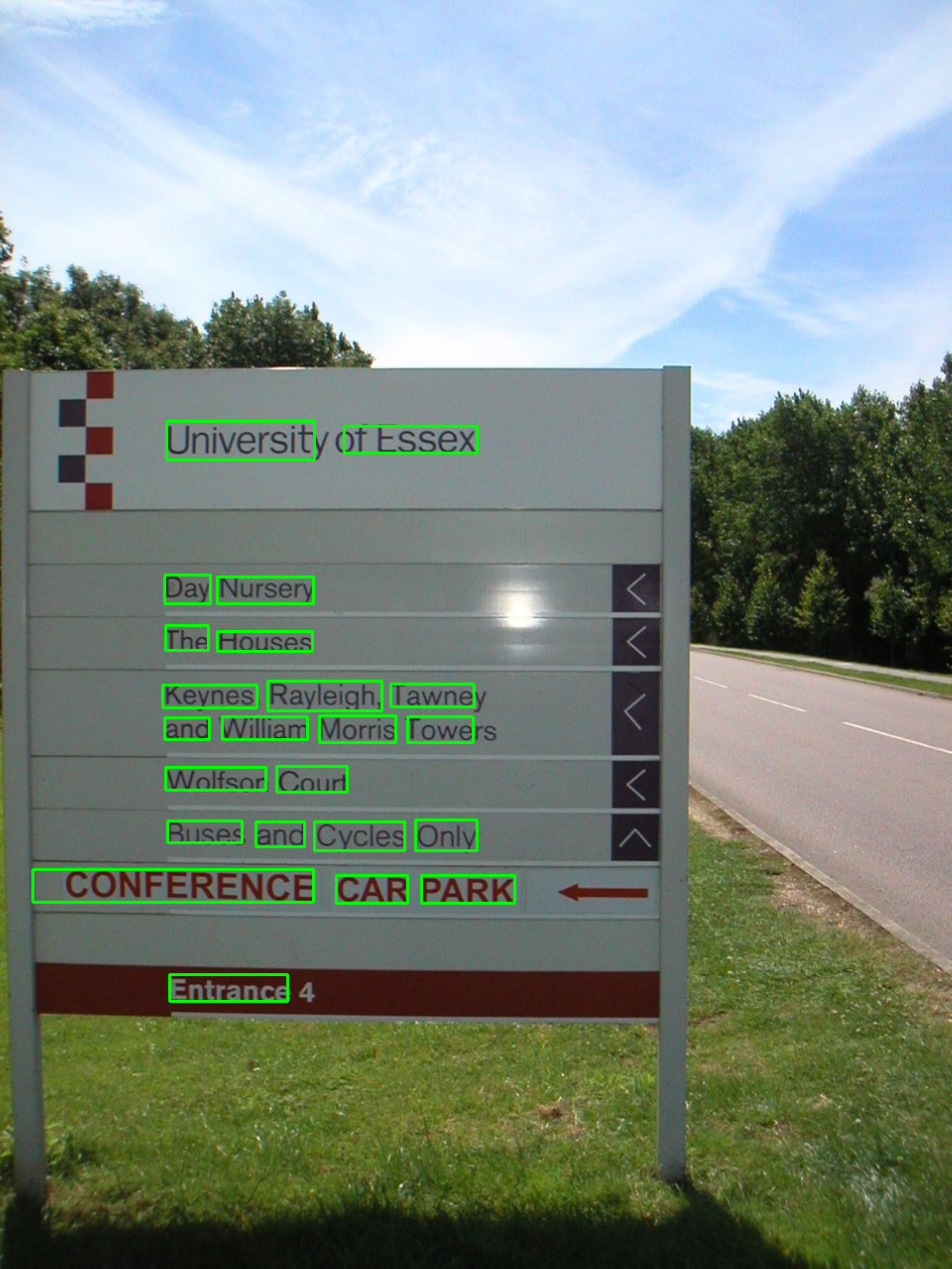}}
	\subfigure[] {
		\label{Fig.sub.icdar13_result_3}
		\includegraphics[width=2.5cm, height=3cm]{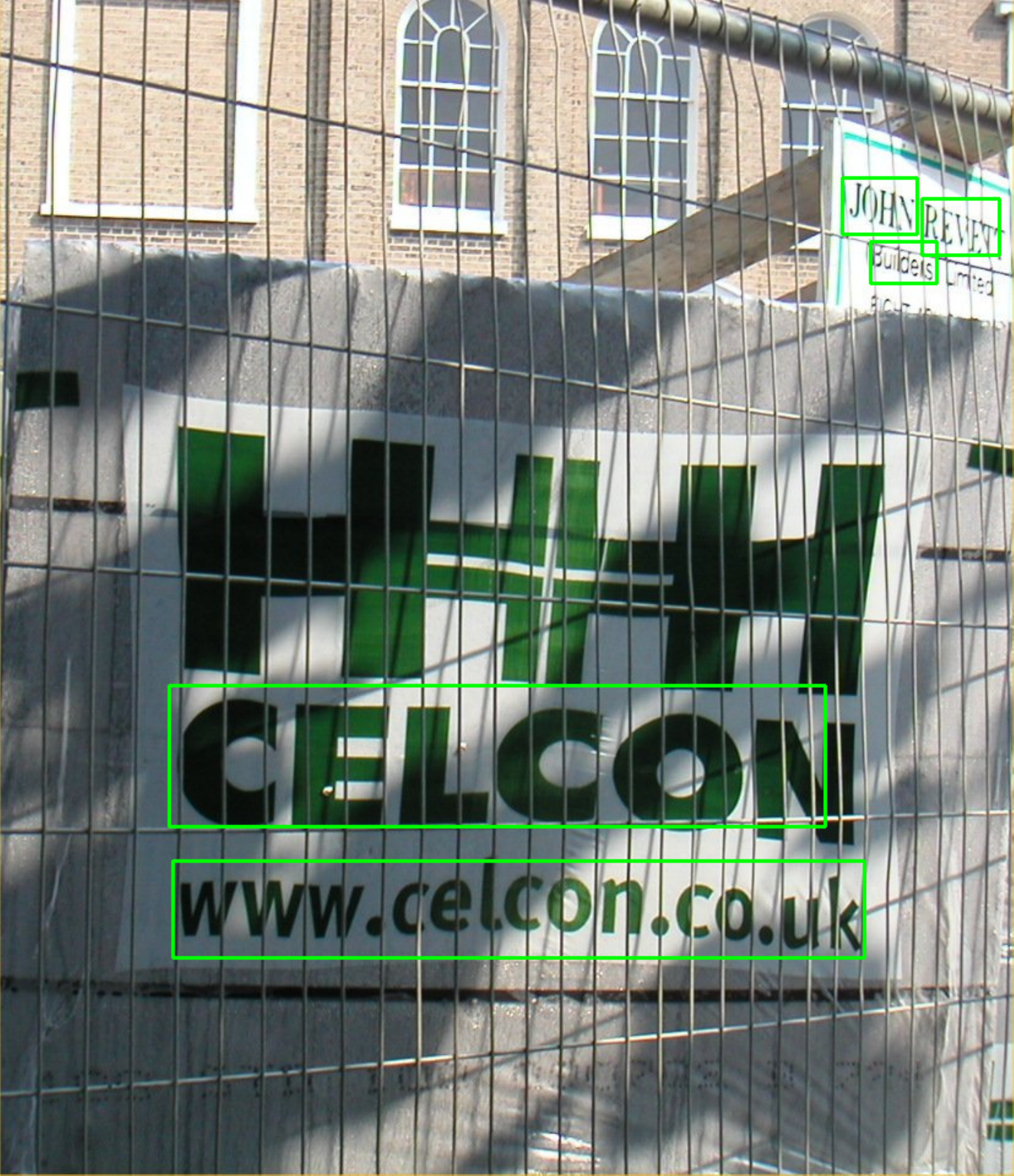}}
	\subfigure[] {
		\label{Fig.sub.icdar13_result_4}
		\includegraphics[width=2.5cm, height=3cm]{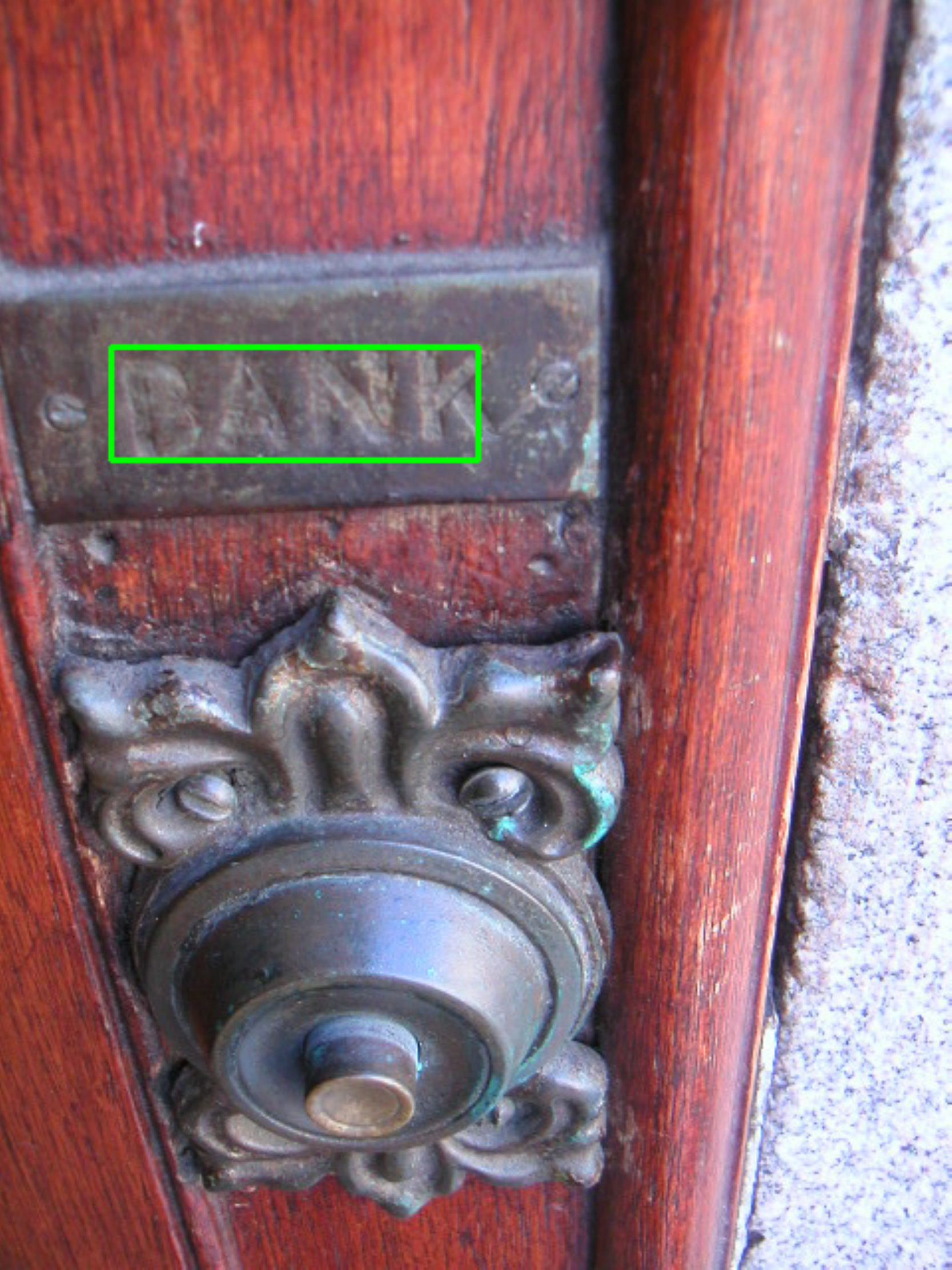}}
	\subfigure[] {
		\label{Fig.sub.icdar13_result_5}
		\includegraphics[width=4cm, height=3cm]{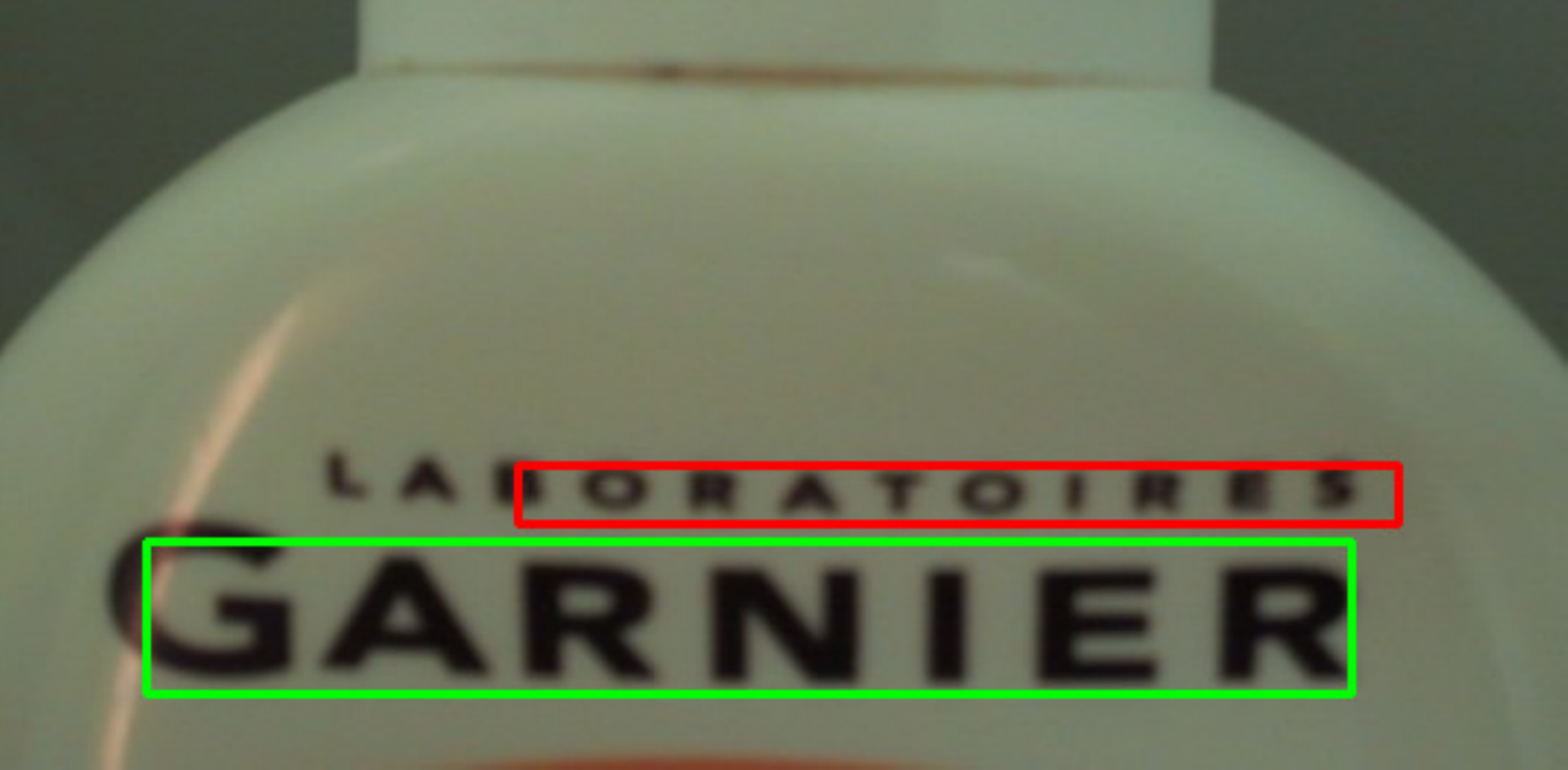}}
	\subfigure[] {
		\label{Fig.sub.icdar13_result_7}
		\includegraphics[width=4cm, height=3cm]{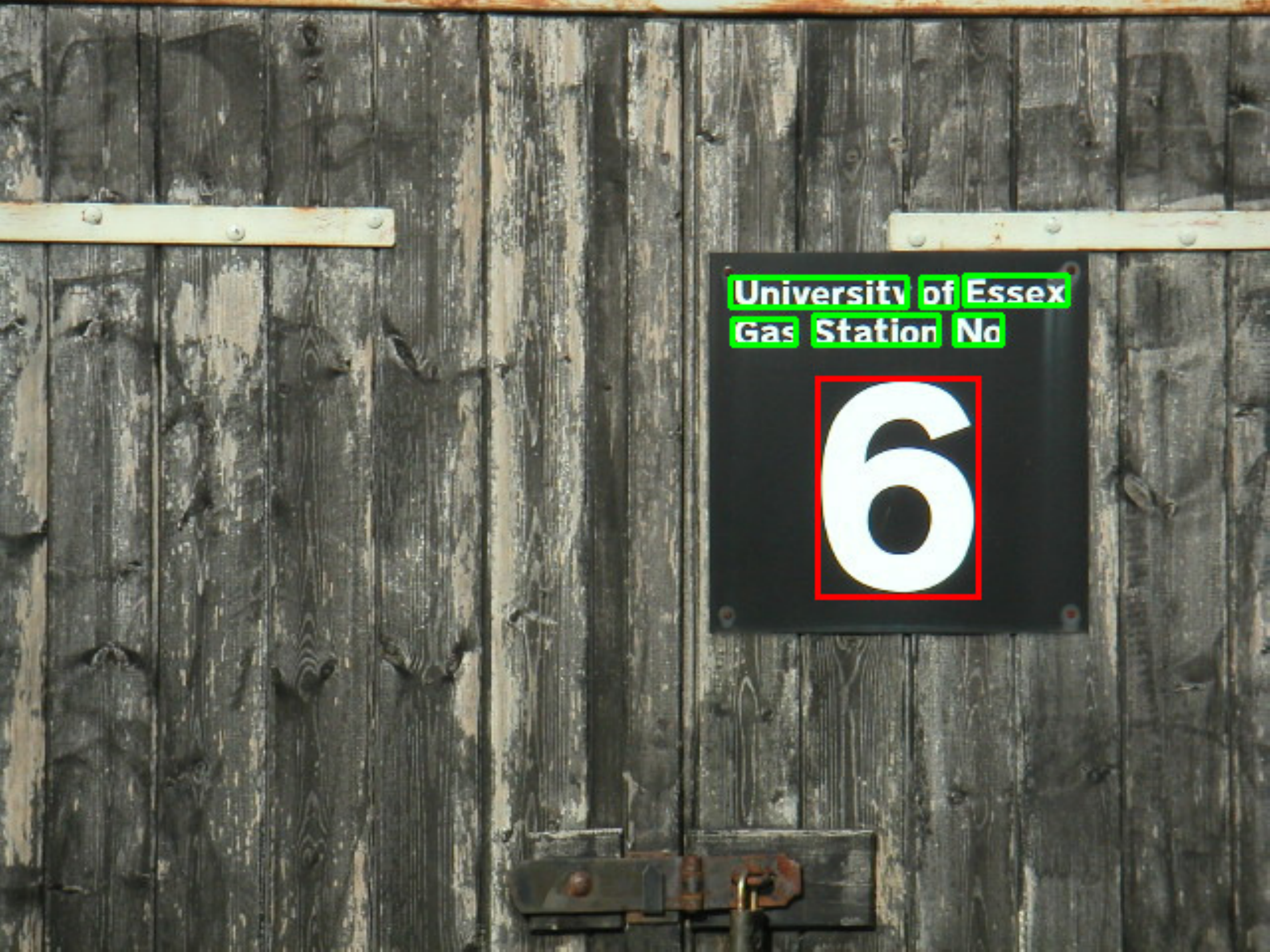}}

	\caption{Detection examples of our model on ICDAR2013. (a)-(c) word level detection for cluttered scenes. (d)-(e) Some failure cases for single character text and losing characters at either end. False and miss detected texts are enclosed by red lines.}
\end{figure}

\noindent \textbf{ICDAR2013 Focused Scene Text.} The detection results of our method on the ICDAR2013 dataset are shown in Tab.\hyperref[Tab.icdar13_result]{3}.
The performance of our method is also the new state-of-the-art. 
Apart from the precision, recall and F-measure, we also list the time cost of our method for per image. 
From the Tab.\hyperref[Tab.icdar13_result]{3} we can see our method is also competitively fast in running speed.
Failed cases are mainly caused by single character text and the inability to enclose letters at either end.
Part of our detection results are shown in Fig.\hyperref[Fig.icdar13_result]{9}.

\begin{table}
	\label{Tab.icdar13_result}
	\small
	\renewcommand\arraystretch{1.2}
	\centering
	\caption{Comparison of methods on ICDAR2013 Focused Scene Text dataset.}
	\begin{tabular}{|c|c|c|c|c|}
		\hline
		Algorithm & Precision & Recall & F-measure & Time \\ 
		\hline
		\hline
		Proposed & 0.92 & 0.81 & \textbf{0.86} & 0.9s \\
		\hline
		Liao \emph{et al.} \cite{textbox} & 0.88 & \textbf{0.83} & 0.85 & \textbf{0.73s} \\
		\hline
		Zhang \emph{et al.} \cite{fcn-text} & 0.88 & 0.78 & 0.83 & 2.1s \\
		\hline
		He \emph{et al.} \cite{he2016text} & \textbf{0.93} & 0.73 & 0.82 & -- \\
		\hline
		Tian \emph{et al.} \cite{textflow} & 0.85 & 0.76 & 0.80 & 1.4s \\
		\hline
	\end{tabular}
\end{table}

\subsection{Rationality of High Performance}
\label{Sec.4.4}
The proposed method is intrinsically able to detect texts of arbitrary orientation, and able to partition words automatically.
The tremendous improvements in both precision and recall for incidental text is mainly attributed to three aspects.

First, direct regression based detection structure avoids to generate proper proposals for irregular shaped multi-oriented texts and thus is more straightforward and effective for multi-oriented scene text detection.

Second, the restriction of positive text size guarantees the robustness of feature representation learned by deep convolutional neural networks. Features for small texts could fade a lot after the first down-sampling operations, and large texts would lose much context information causing the CNN could only see some simple strokes of the large texts. Texts within a proper scale range could contain both text textures and enough semantic context making the CNN learn more robust scene text features.
Moreover, the classification task which is able to distinguish text and non-text regions providing a solid foundation for regression task.

Third, the end-to-end optimization mechanism to localize text is much more robust than rule based methods.
Previous methods treating line grouping and word partition as post processing are prone to lose much useful information and rely on thresholds chosen, but integrating localization into the network for end-to-end training could well solve the mentioned issues above.

\subsection{Comparison to Other Scene Text Detection Systems}
\label{Sec.4.5}

Here we list and compare with some recent high performance scene text detection methods for better understanding on the superiority of our method. The listed methods are arranged by the time they are proposed.

\noindent \textbf{TextFlow.} TextFlow \cite{textflow} is designed for horizontal scene text detection by extracting character candidates firstly and then group characters into text lines. Its main contribution is to reduce the traditional multi-module system into fewer steps. Due to the more integrated pipeline, it could reach competitive performance for horizontal text detection.
We take benefits of its intuition and design a simpler process to detect text words/lines directly without extracting character candidates or line grouping.

\noindent \textbf{SymmetryText.} SymmetryText \cite{symmetry-text} might be the first work that treats scene text detection as object detection. It proposes symmetric feature and uses it to generate text line proposals directly. However, the symmetric feature is not robust for cluttered scenes or adaptive to multi-oriented text. In our work, we skip the text line proposal generation step and adopt the deep convolutional feature which is more robust and representative.

\noindent \textbf{FCNText.} FCNText \cite{fcn-text} adopts the FCN \cite{fcn} for object segmentation to segment the text region by a coarse-to-fine process. The employment of deep convolutional features ensures accurate localization of text regions. To output the bounding box for each text word/line, FCNText resorts to some heuristic rules to combine characters into groups. In our work, we abandon the character-to-line procedure to get a more straightforward system and less parameters for tuning.

\noindent \textbf{FCRN.} FCRN \cite{yolo-text} is modified from YOLO for scene text detection. Both FCRN and YOLO perform bounding box regression much like direct regression, however, they actually adopt a compromise strategy between direct and indirect regression for they use multiple non-predefined candidate boxes for direct regression, and hopes candidate boxes behave like anchors in \cite{faster-rcnn} after well optimized.
Another important difference between FCRN and our method is that both FCRN and YOLO regard the centroid region as positive, while we regard regions around the text center line as positive. Our definition of positive/text region seems more proper since text features are alike along the text center line.

\noindent \textbf{CTPN.} CTPN \cite{ctpn} can be deemed as an upgraded character-to-line scene text detection pipeline. It first adopts the RPN in Faster-RCNN to detect text slices rather than characters within the text regions and then group these slices into text bounding boxes. The text slices could be more easily integrated into an end-to-end training system than characters and more robust to represent part of the text regions. In our work, we follow a different way by detecting the whole texts rather than part of the texts.

\noindent \textbf{TextBoxes \& DeepText.} TextBoxes \cite{textbox} and DeepText \cite{deep-text} are based on SSD and Faster-RCNN respectively. They both take advantages from the high performance object detection systems and treat text word/line as a kind of generic object. Moreover, they both set anchors to have more varieties and can only detect horizontal scene texts. In our work, we perform the regression by a direct way and can tackle with multi-oriented text detection.

\noindent \textbf{DMPN.} DMPN \cite{dmpn} is an indirect regression based method and it also treats text detection as object detection. Unlike TextBoxes or DeepText, it introduces a multi-oriented anchor strategy to find the best matched proposal in parallelogram form to the arbitrary quadrilateral boundaries of multi-oriented texts. However, as \cite{dmpn} itself refers, DMPN relies on the man-made shape of anchors which may not be the optimal design and this fits well with our analysis on the drawbacks of indirect regression. The large margin of performance between DMPN and our method on ICDAR2015 Incidental Text benchmark also verify the significance of our work.

\section{Conclusion}
\label{Sec.5}
In this paper, we first partition existing object detection frameworks into direct and indirect regression based methods, and analyze the pros and cons of both methods for irregular shaped object detection.
Then we propose a novel direct regression based method for multi-oriented scene text detection. 
Our detection framework is straightforward and effective with only one-step post processing. Moreover it performs particularly well for incidental text detection.
On the ICDAR2015 Incidental Scene Text benchmark, we have achieved a new state-of-the-art performance and outperformed previous methods by a large margin.
Apart from this, we also analyze the reasons of the high performance and compare our method to other recent scene text detection systems.
Future work will focus on more robust and faster detection structure, as well as more theoretical research on regression task.

{\small
\bibliographystyle{ieee}
\bibliography{egbib}
}

\end{document}